\documentclass[runningheads]{llncs}

\usepackage[utf8]{inputenc}
\DeclareUnicodeCharacter{2194}{\ensuremath{\leftrightarrow}}
\usepackage{eccv}

\usepackage{eccvabbrv}

\usepackage{graphicx}
\usepackage[export]{adjustbox}
\usepackage{booktabs}
\usepackage{multirow}
\usepackage{caption}
\usepackage{enumitem}
\usepackage{wrapfig}
\usepackage{colortbl}

\usepackage[accsupp]{axessibility}  

\usepackage{hyperref}

\usepackage{orcidlink}

\usepackage{xcolor}

\graphicspath{{figures/}}

\newcommand{\shortname}{LumiTokens\xspace}

\definecolor{myorange}{RGB}{255, 140, 0}

\begin{document}

\title{LumiTokens: 3D Relighting via Token-Space Lighting Transformation} 

\titlerunning{Abbreviated paper title}

\author{Yiwen Chen\inst{1} \and
Matheus Gadelha\inst{2} \and
Huaizu Jiang\inst{1}}

\authorrunning{Y.~Chen et al.}
\titlerunning{LumiTokens}

\institute{Northeastern University \and
Adobe Research\\
\url{https://neu-vi.github.io/LumiTokens}
}

\maketitle

\begin{figure}[!h]
    \centering
    \setlength{\tabcolsep}{0pt}
    \renewcommand{\arraystretch}{1.0}

    \begin{tabular}{@{}
        c @{\hspace{0.035\linewidth}}
        c @{\hspace{0.006\linewidth}} c @{\hspace{0.035\linewidth}}
        c @{\hspace{0.006\linewidth}} c @{}}
        \multicolumn{1}{c}{\footnotesize\sffamily\bfseries Sparse-view inputs} &
        \multicolumn{2}{c}{\footnotesize\sffamily\bfseries Relit under an envmap} &
        \multicolumn{2}{c}{\footnotesize\sffamily\bfseries Relit under a point light} \\[2pt]
        \includegraphics[height=0.175\linewidth]{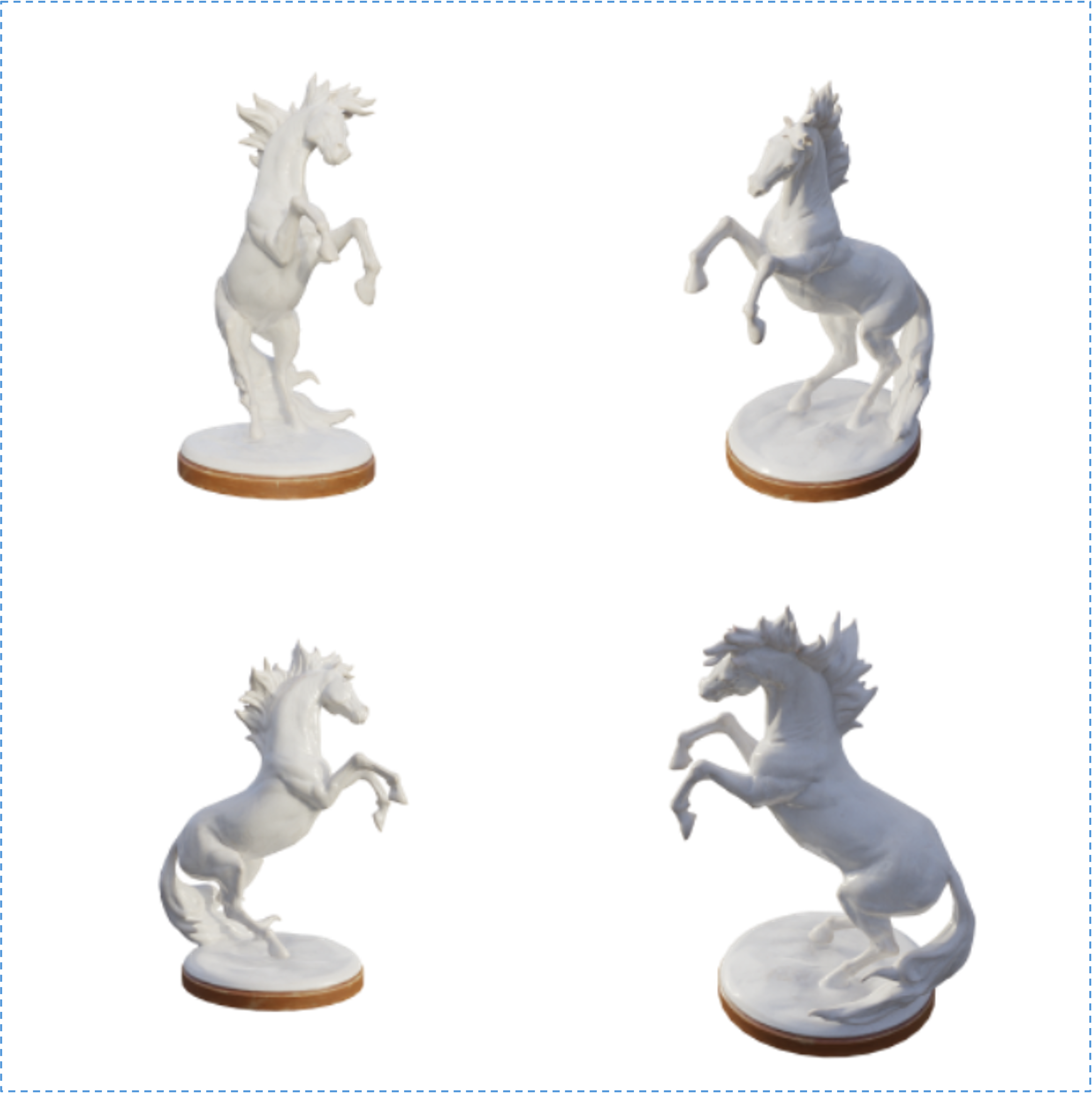} &
        \includegraphics[height=0.175\linewidth]{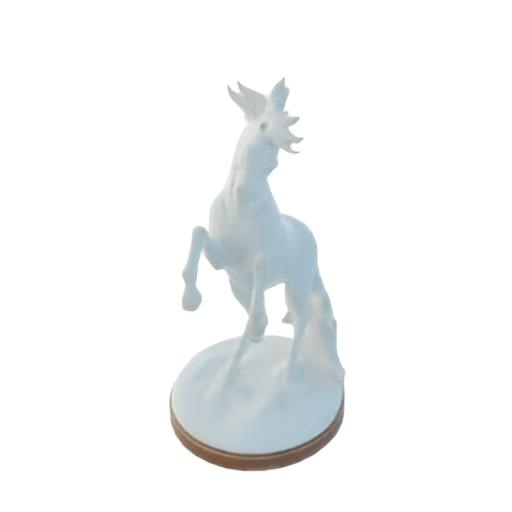} &
        \includegraphics[height=0.175\linewidth]{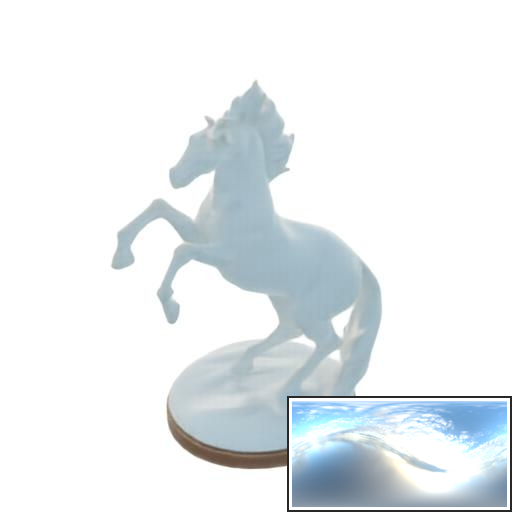} &
        \includegraphics[height=0.175\linewidth]{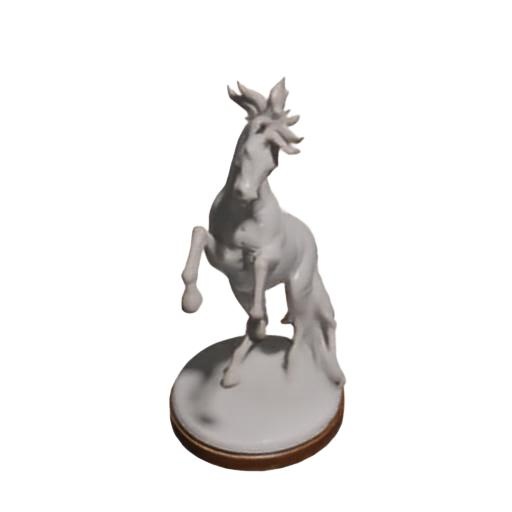} &
        \includegraphics[height=0.175\linewidth]{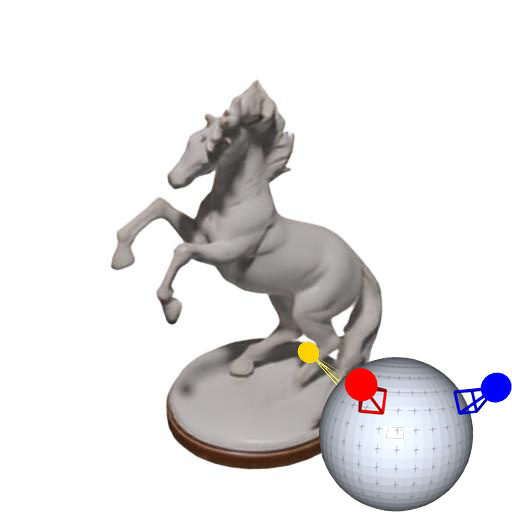} \\[2pt]
        \includegraphics[height=0.175\linewidth]{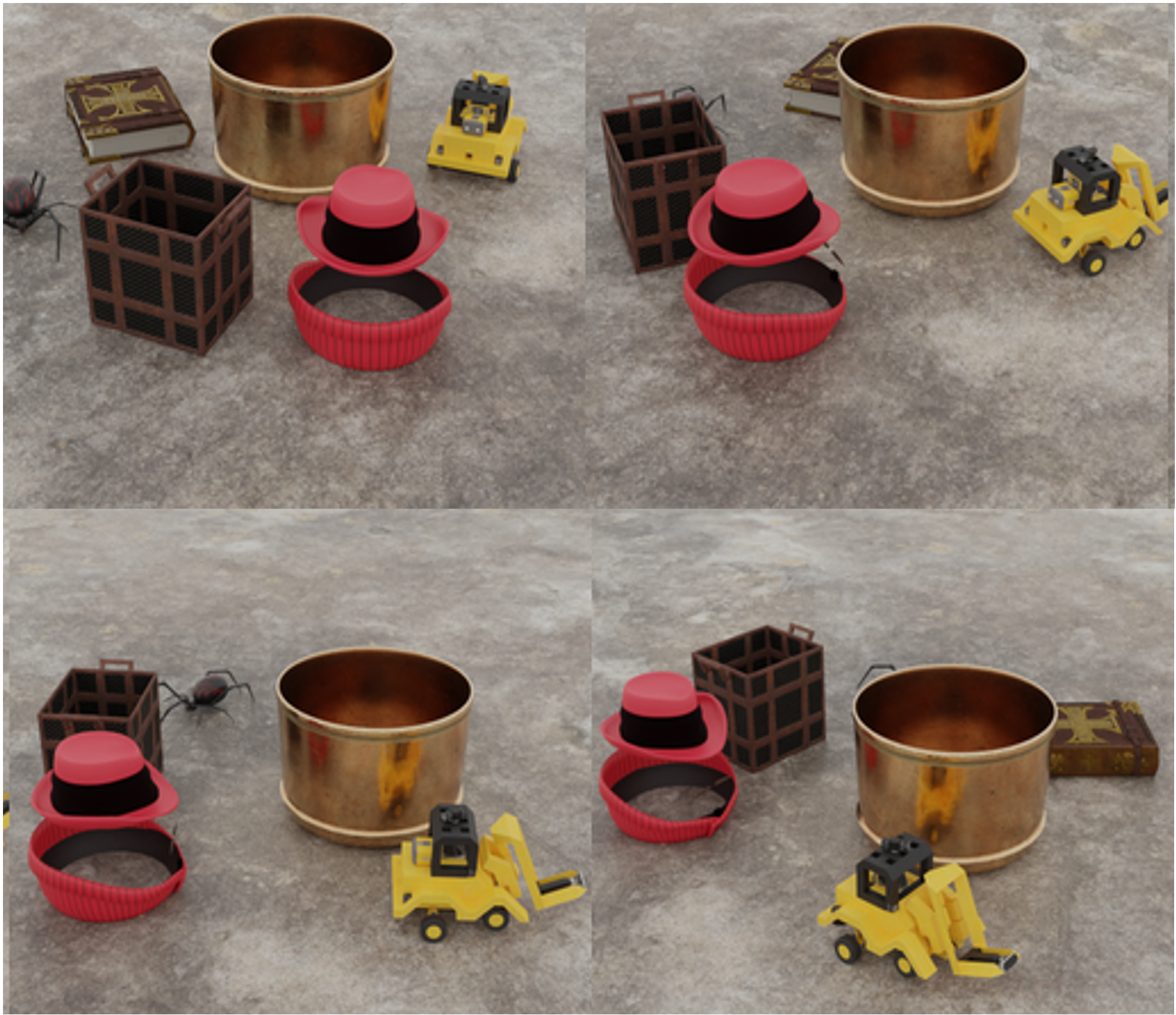} &
        \includegraphics[height=0.175\linewidth]{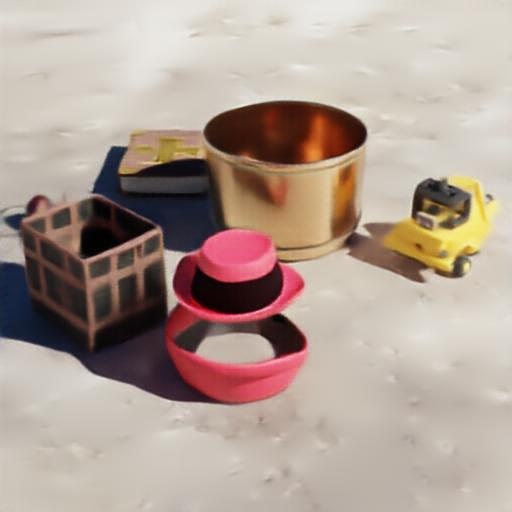} &
        \includegraphics[height=0.175\linewidth]{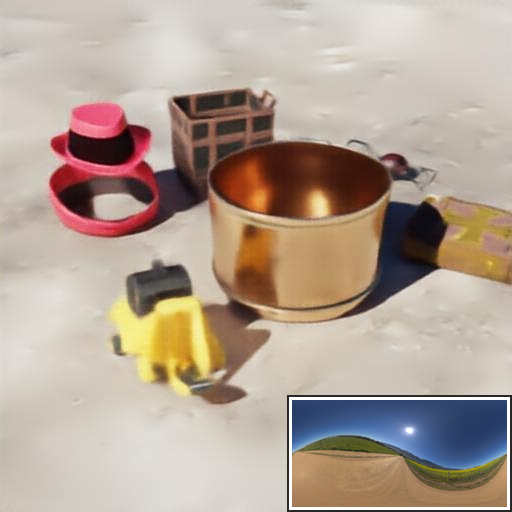} &
        \includegraphics[height=0.175\linewidth]{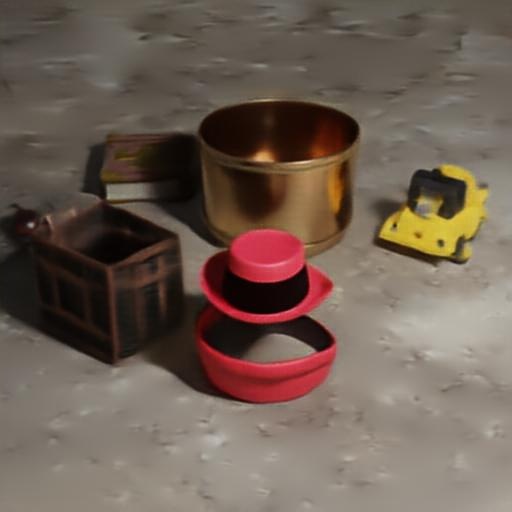} &
        \includegraphics[height=0.175\linewidth]{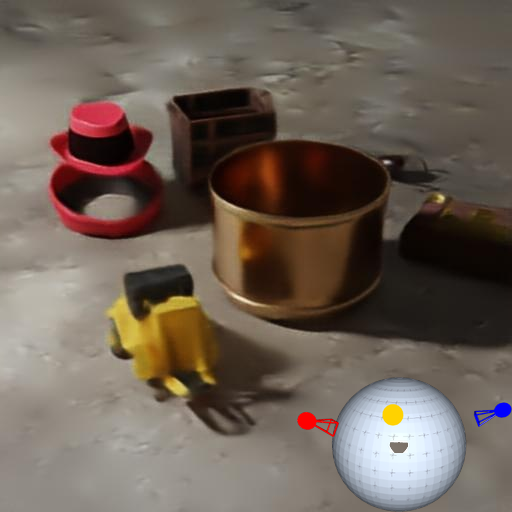} \\
    \end{tabular}

    \noindent\textcolor{black!40}{\rule{\linewidth}{0.4pt}}

    \begin{tabular}{@{}
        c @{\hspace{0.012\linewidth}} c @{\hspace{0.012\linewidth}}
        c @{\hspace{0.012\linewidth}} c @{\hspace{0.012\linewidth}} c @{}}
        \multicolumn{5}{c}{\small\sffamily\bfseries Progressively relighting a scene} \\[2pt]
        \includegraphics[width=0.188\linewidth]{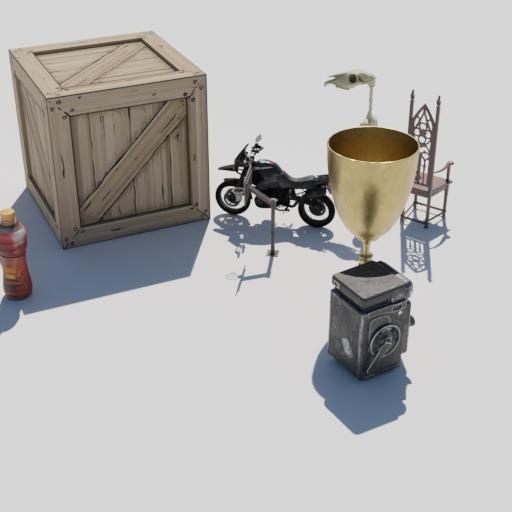} &
        \includegraphics[width=0.188\linewidth]{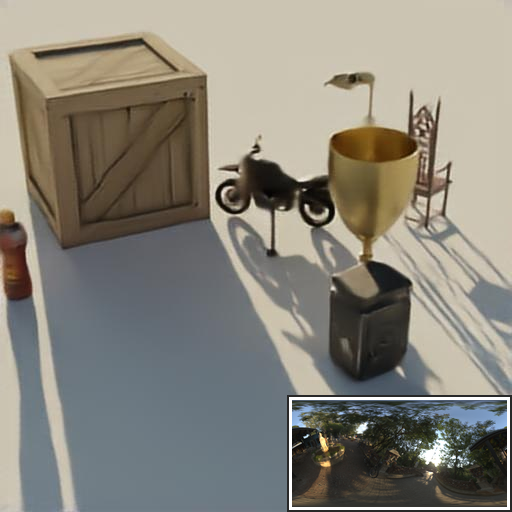} &
        \includegraphics[width=0.188\linewidth]{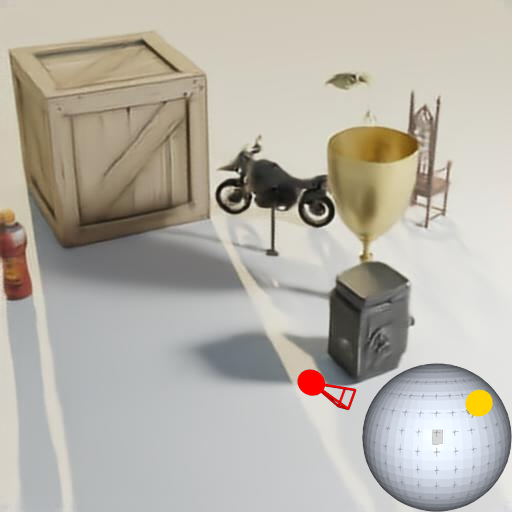} &
        \includegraphics[width=0.188\linewidth]{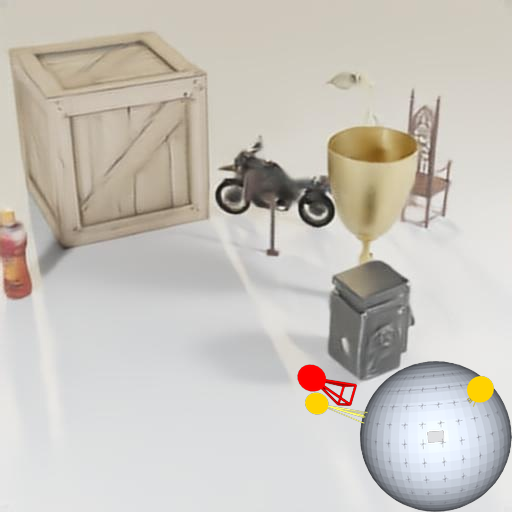} &
        \includegraphics[width=0.188\linewidth]{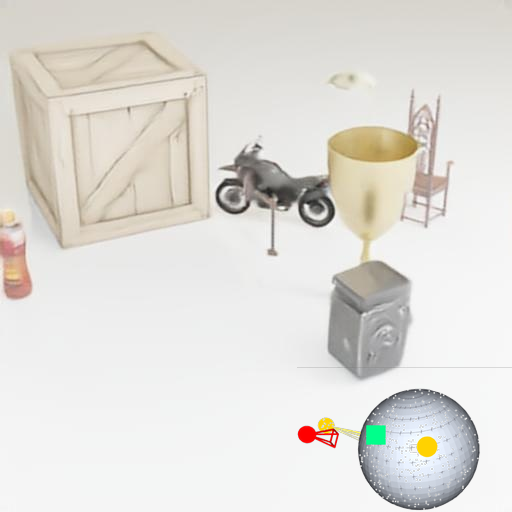} \\[1pt]
        \parbox[t]{0.188\linewidth}{\centering\tiny\sffamily input} &
        \parbox[t]{0.188\linewidth}{\centering\tiny\sffamily under a new env light} &
        \parbox[t]{0.188\linewidth}{\centering\tiny\sffamily add a pointlight} &
        \parbox[t]{0.188\linewidth}{\centering\tiny\sffamily add another pointlight} &
        \parbox[t]{0.188\linewidth}{\centering\tiny\sffamily add an area light} \\
    \end{tabular}

    \caption{Given multi-view input images (left), \textbf{LumiTokens} relights the scene under diverse lighting types.
\textit{Top:} an object relit under an environment map and a point light, rendered from novel views.
The red and blue dots indicate the rendering views, yellow dots show the position of the point light sources in 3D, and the cyan rectangle illustrates an area light.
\textit{Middle:} relighting generalizes to multi-object scenes with inter-object shadows and reflections.
\textit{Bottom:} progressive relighting, where the user incrementally adds light sources to edit the lighting condition.}
    \label{fig:teaser}
\end{figure}

\begin{abstract}
Existing 3D relighting methods operate through either explicit material decomposition, diffusion-based view-space generation, or a combination of both, requiring full recomputation for each new lighting condition.
We observe that recent latent scene representations, which encode multi-view images into a set of compact tokens with no fixed physical semantics, open up a novel design space for relighting.
We present \textbf{LumiTokens}, a framework that formulates 3D relighting as a direct transformation on latent scene tokens, without explicit 3D representations, rendering equations, or physics-based decomposition.
Our model introduces a Scene Token Editor that processes scene tokens jointly with light-ray tokens through self-attention, producing updated tokens that can be decoded into multi-view-consistent relit images.
To support diverse lighting types through a unified interface, all lighting signals, including environment maps, point lights, and area lights, are parameterized as Pl\"ucker ray tokens, enabling native 3D user interaction with a representation that carries no explicit spatial structure.
Crucially, this design supports \emph{progressive relighting}: because the editor's output remains in the same latent space as its input, a user can incrementally build up illumination one light source at a time, with each edit composing in token space.
Experiments demonstrate that LumiTokens achieves comparable or superior relighting quality to other methods
and supports progressive, composable lighting edits.
\keywords{3D Relighting \and Latent Scene Tokens \and Progressive Relighting}
\end{abstract}
\section{Introduction}

3D scene reconstruction from multi-view captures has seen remarkable progress.
Methods such as Neural Radiance Fields (NeRF)~\cite{NeRF} and 3D Gaussian Splatting (3DGS)~\cite{3DGS} can reconstruct complex objects and scenes with high fidelity, and are increasingly used to create 3D assets for film production, video games, and e-commerce.
For these assets to be deployed in new environments, however, they must be rendered under novel illumination: a character scanned in a studio needs to match the on-set lighting of a film, a product captured in a lab must look natural in a customer's living room, and a virtual object in augmented reality should cast shadows consistent with the real world.
Relighting, the ability to render a reconstructed scene under arbitrary target illumination, is therefore a critical step for bridging 3D reconstruction and practical downstream applications.

\begin{figure}[t]
    \centering
    \includegraphics[width=\linewidth]{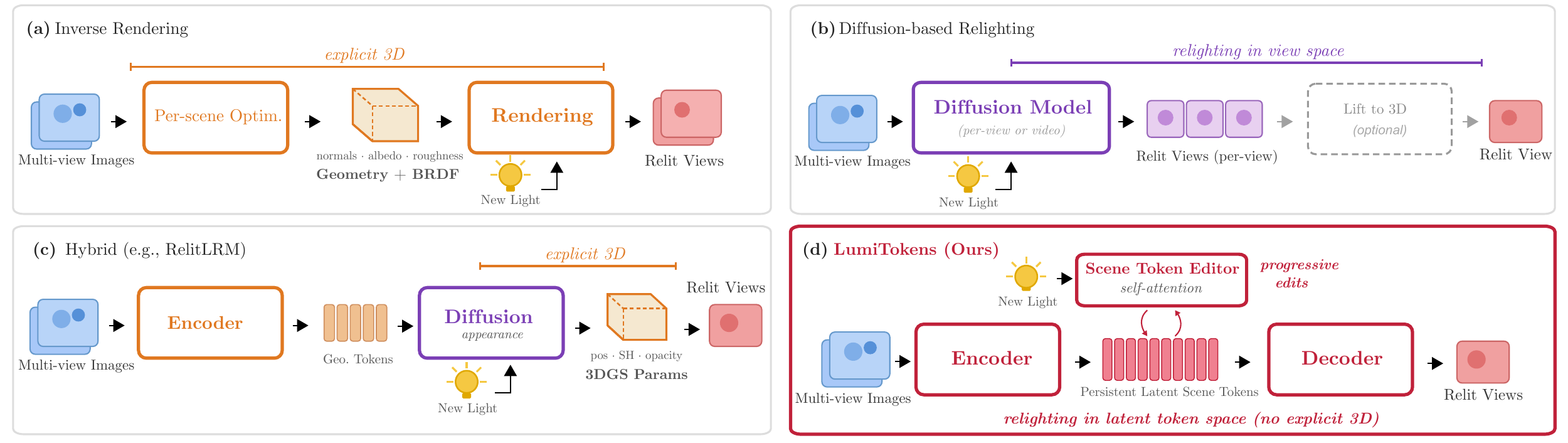}
    \caption{
   \textbf{Comparison of relighting paradigms}.
   (a) Inverse rendering recovers explicit geometry and BRDF properties via per-scene optimization, limited by the expressiveness of the material model.
   (b) \textcolor{black}{Diffusion-based methods use powerful generative priors to synthesize relit views, without a shared scene representation to anchor cross-view consistency or persist lighting edits.}
   (c) Hybrid methods combine feed-forward geometry encoding with generative appearance synthesis, inheriting limitations from both paradigms.
   (d) LumiTokens (Ours) performs relighting entirely in latent token space: scene tokens are edited via self-attention with light-ray tokens and decoded into relit views, supporting progressive, incremental lighting edits. 
    }
    \label{fig:paradigm_comparison}
\end{figure}

To tackle the relighting problem, existing methods can be grouped into three categories (Fig.~\ref{fig:paradigm_comparison}), each with fundamental tradeoffs.
The first relies on \textbf{inverse rendering with explicit 3D representations}.
These methods recover intrinsic material properties such as surface normals, albedo, and roughness from multi-view images, and relight by re-evaluating a rendering equation under novel illumination~\cite{shape_illumination_reflectance_from_shading,NeRD,phySG,NerFactor,i2sdf,diffusion_intrinsic_estimation,TensoIR,Relightable3DGS}.
While inherently multi-view consistent due to their explicit 3D structure, these approaches typically require dense multi-view captures and costly per-scene optimization.
They are further limited by the expressiveness of their chosen BRDF model and rendering equation.

The second family leverages \textbf{diffusion models} as powerful appearance priors for relighting, bypassing explicit material decomposition and producing realistic lighting effects across diverse materials~\cite{dilightNet,neuralGaffer,ICLight,rgbx,diffusion_intrinsic_estimation}.
While early methods operate on single images, recent extensions adopt video or multi-view diffusion architectures that encourage cross-view coherence through temporal priors or multi-view attention mechanisms~\cite{DiffusionRenderer,GenLit,UniRelight,LightSwitch,X2Video,CtrlVDiff,DiffusionRadianceFieldRelighting,IllumiNeRF,GR3EN,LuxRemix,ROGR}.
Despite their strong results, these methods primarily formulate relighting as image-, video-, or view-space generation: the edited illumination is not stored as a persistent scene-level representation, and each lighting change typically requires a new generative pass rather than an update to a reusable scene state.

A third, \textbf{hybrid} category has emerged that draws on both paradigms.
These methods build on feed-forward reconstruction models~\cite{LRM,GSLRM}, which use transformers trained on massive 3D datasets to predict explicit 3DGS from sparse views in a forward pass.
To extend this pipeline to relighting, they incorporate diffusion or generative models that produce lighting-dependent appearance conditioned on a target environment map~\cite{RelitLRM,NeAR}.
This combination enables feed-forward relightable reconstruction without per-scene optimization, but the final output remains an explicit 3DGS representation, inheriting the expressiveness ceiling of the 3D parameterization from the first category while still relying on generative appearance synthesis from the second.
Across all three categories, relighting is treated as either a decomposition problem, a view-generation problem, or a combination of both.
In each case, the lighting edit is not maintained as an incremental update to a persistent scene representation.

Recent advances in novel view synthesis have shown that multi-view images can be compressed into a set of compact tokens that serve as fully learned scene representations, without imposing any explicit 3D structure~\cite{SRT,LVSM,RayZer,SceneTok}.
Because these tokens encode an entire scene, including geometry and appearance, jointly in a latent space with no fixed physical semantics, we observe that they open up a novel design space for relighting.
We propose \textbf{relighting as a direct transformation of a latent scene representation conditioned on lighting}.
This stands in contrast to all three existing categories: unlike inverse rendering and hybrid methods, our representation is not bound to explicit 3D parameterizations such as 3DGS; \textcolor{black}{unlike prior diffusion-based relighting methods that use generation as the primary mechanism for producing relit views, our relighting operation updates a shared 3D scene representation that naturally anchors cross-view consistency}; and unlike all existing paradigms, our approach supports \emph{persistent} editing, where lighting changes accumulate in token space without requiring re-evaluation, regeneration, or re-encoding at each step.
To our knowledge, this idea has not been explored in prior work.
Realizing it, however, is non-trivial: the latent space is shaped by a reconstruction objective, and it is unclear whether coherent lighting transformations exist as learnable functions in this space.
Since geometry and appearance are entangled in the tokens, it is uncertain whether one can change illumination without corrupting the encoded scene structure.
And since the tokens carry no built-in 3D structure, it is not obvious how the model can support native 3D user interaction, \eg, responding correctly when a user places a point light at a specific 3D location.

We present \textbf{LumiTokens}, a transformer-based framework for relightable novel view synthesis that operates entirely within a learned latent token space, free of explicit 3D representations and rendering equations.
Our model adopts a three-part architecture: an \emph{encoder} that compresses a variable number of posed multi-view images (from as few as 4 to as many as 32) into a compact set of 1D scene tokens, a \emph{Scene Token Editor} that transforms these tokens conditioned on target illumination, and a \emph{decoder} that renders the modified scene from novel viewpoints.
To support diverse lighting types through a unified interface, we parameterize all lighting signals as Pl\"ucker ray tokens: environment maps, point lights, and area lights are all discretized into light rays, enabling native 3D user interaction with a representation that carries no explicit spatial structure.
The editor processes scene tokens and light-ray tokens jointly through self-attention, producing updated tokens that the decoder renders into correctly relit novel views.
A key design requirement is that the editor's output remains in the same latent space as its input, making the scene tokens a \emph{persistent, editable representation}: a user can incrementally build up the illumination one source at a time, with the editing chain staying in token space and the decoder serving as a read-only preview at any step.

To summarize, our contributions are as follows:
\begin{itemize}[itemsep=0pt,topsep=0pt,parsep=0pt,partopsep=0pt]
    \item We demonstrate that 3D relighting can be performed as a direct transformation in a learned latent token space, without explicit 3D representations, rendering equations, or physics-based decomposition.
    \item We introduce the Scene Token Editor, a transformer module that performs relighting through self-attention between scene tokens and light-ray tokens encoded as Pl\"ucker rays, supporting environment maps, point lights, and area lights through a unified interface and enabling native 3D user interaction.
    \item We show that our formulation supports progressive relighting: because the editor's output remains in the same latent space as its input, lighting edits compose in token space, enabling incremental lighting design that existing paradigms can achieve only through full recomputation at each step.
\end{itemize}
\section{Related Works}

\noindent\textbf{Latent Scene Representations.}
A growing body of work has shown that 3D scenes can be effectively represented as compact sequences of learned tokens, without imposing explicit 3D structure such as NeRF~\cite{NeRF}, 3DGS~\cite{3DGS}, or triplanes~\cite{EG3D}.
SRT~\cite{SRT} introduced this paradigm by encoding multi-view images into set-latent scene representations from which novel views are decoded via cross-attention.
LVSM~\cite{LVSM} scaled this idea to a large transformer architecture with minimal 3D inductive bias, demonstrating that an encoder-decoder model operating on 1D latent tokens can match or surpass explicit 3DGS-based methods such as GS-LRM~\cite{GSLRM} on novel view synthesis, particularly on challenging materials involving specular and transparent surfaces.
RayZer~\cite{RayZer} and its successor E-RayZer~\cite{ERayZer} further showed that such representations can be learned in a fully self-supervised manner without camera pose annotations.
SceneTok~\cite{SceneTok} demonstrated that unstructured, permutation-invariant tokens can achieve orders-of-magnitude compression of 3D scenes while remaining compatible with generative models.
These methods have been used exclusively for novel view synthesis.
Extending these representations from passive encoding/decoding to active manipulation for relighting is non-trivial, as discussed in Sec.~\ref{subsec:latent_scene}.
Our work is the first to show that the same class of latent scene tokens can support relighting through a direct, composable transformation in token space.


\noindent\textbf{Inverse Rendering for Relighting.}
The classical approach to relighting formulates it as an inverse rendering problem: recover intrinsic scene properties (geometry, materials, lighting) from multi-view images, then re-render under novel illumination via a physically based rendering equation~\cite{shape_illumination_reflectance_from_shading}.
Early neural methods such as NeRD~\cite{NeRD}, PhySG~\cite{phySG}, and NeRFactor~\cite{NerFactor} extended this paradigm to NeRF-based representations, learning spatially varying BRDF parameters alongside the radiance field.
Subsequent work improved efficiency and physical fidelity: TensoIR~\cite{TensoIR} achieved significant speedups through tensor-factorized representations with online visibility computation, I$^2$-SDF~\cite{i2sdf} extended inverse rendering to indoor scenes with near-field lighting, and Relightable 3D Gaussians~\cite{Relightable3DGS} brought the paradigm to 3DGS with real-time rendering via BVH-based ray tracing.
Other methods incorporate learned priors to improve robustness, such as pre-trained depth estimators~\cite{NeROIC} or diffusion-based material prediction~\cite{diffusion_intrinsic_estimation,GS-IR}.
Despite steady progress, these methods share two fundamental limitations: they are bounded by the expressiveness of their chosen BRDF model, which cannot capture complex effects such as subsurface scattering or inter-reflections, and most require dense multi-view captures with costly per-scene optimization.

\noindent\textbf{Diffusion-based Relighting.}
An alternative line of work bypasses explicit material decomposition, instead leveraging the strong priors of diffusion models trained on large-scale data to generate realistic lighting effects.
Single-image methods such as DiLightNet~\cite{dilightNet}, Neural Gaffer~\cite{neuralGaffer}, IC-Light~\cite{ICLight}, and RGB$\leftrightarrow$X~\cite{rgbx} demonstrated that diffusion models can produce high-quality relightings conditioned on environment maps or lighting descriptors, but each image is relit independently with no multi-view consistency.
To address this, recent extensions introduce 3D-awareness through various mechanisms.
IllumiNeRF~\cite{IllumiNeRF} relights individual views with a diffusion model and distills the results into a Latent NeRF via per-scene optimization.
DiffusionRenderer~\cite{DiffusionRenderer} uses video diffusion priors with cross-attention injection for temporal consistency.
LightSwitch~\cite{LightSwitch} enforces multi-view coherence through multi-view self-attention in the denoising UNet, while UniRelight~\cite{UniRelight} jointly models albedo estimation and relighting in a single Diffusion Transformer pass.
GenLit~\cite{GenLit} reformulates relighting as video generation, and ROGR~\cite{ROGR} trains a lighting-conditioned NeRF from diffusion-generated relit images.
At the scene level, GR3EN~\cite{GR3EN} and LuxRemix~\cite{LuxRemix} extend generative relighting to room-scale environments with per-luminaire control.
Other works explore controllable video diffusion for relighting as part of broader multimodal editing frameworks~\cite{X2Video,CtrlVDiff,DiffusionRadianceFieldRelighting}.
Despite impressive quality, these methods lack a shared scene representation to anchor cross-view consistency or persist lighting edits, and each new lighting condition requires a full generative pass.


\noindent\textbf{Hybrid and Feed-forward Relighting.}
Feed-forward large reconstruction models such as LRM~\cite{LRM} and GS-LRM~\cite{GSLRM} predict explicit 3D representations (triplane NeRF and 3DGS, respectively) from sparse views in a single forward pass.
Several methods extend this backbone to support relighting: RelitLRM~\cite{RelitLRM} adds a diffusion-based appearance generator alongside a deterministic geometry branch, NeAR~\cite{NeAR} uses a rectified-flow model to produce lighting-homogenized representations decoded into relightable 3DGS, and GRGS~\cite{GRGS} predicts explicit material properties on Gaussians for human relighting.
However, their output is an explicit 3DGS rendered through splatting, and each new lighting condition requires regenerating appearance through the diffusion or generative branch.
Our approach differs in that relighting operates entirely within a latent token space with no explicit 3D output, and edited tokens persist across lighting changes without regeneration (Sec.~\ref{subsec:latent_scene}).



\section{Method}

\begin{figure}[t]
    \centering
    \includegraphics[width=\linewidth]{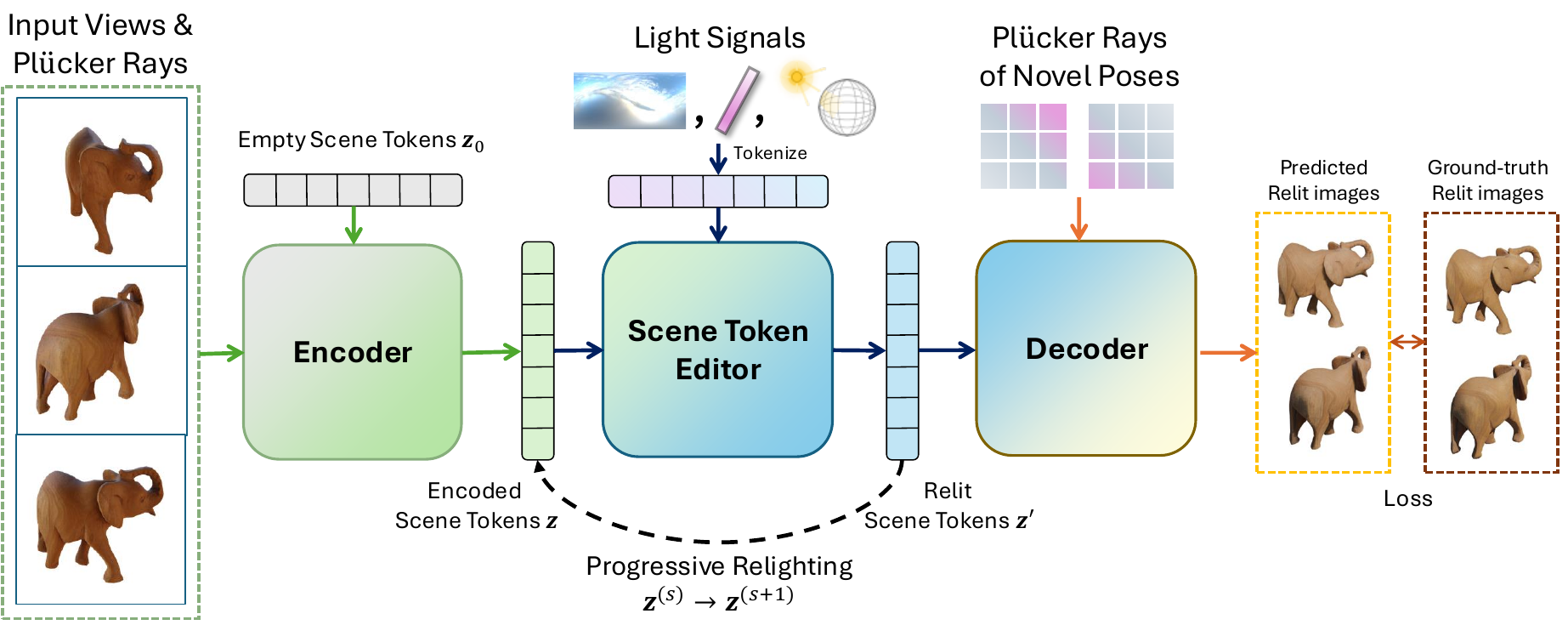}
    \caption{\textbf{Overview of the LumiTokens architecture.} An encoder maps posed multi-view images, together with learnable empty tokens $\mathbf{z}_0$, into unstructured latent scene tokens $\mathbf{z}$. The scene token editor transforms these tokens conditioned on target lighting: all light types (environment maps, point lights, area lights) are discretized into Pl\"ucker ray tokens and processed jointly with the scene tokens through self-attention, producing relit scene tokens $\mathbf{z}'$. A decoder with a DPT head then renders $\mathbf{z}'$ into novel views given target camera Pl\"ucker rays. Because the editor's output remains in the same latent space as its input, the architecture naturally supports progressive relighting (dashed arrow): multiple lighting edits chain directly in token space without decoding between steps.}
    \label{fig:architecture}
\end{figure}

\subsection{Latent Scene Representations for Relighting}
\label{subsec:latent_scene}

Recent work on novel view synthesis has shown that multi-view images can be compressed into compact, \emph{unstructured} token sequences that serve as fully learned scene representations~\cite{SRT,LVSM,RayZer,SceneTok}.
Unlike explicit 3D representations such as NeRF~\cite{NeRF} or 3DGS~\cite{3DGS}, these methods impose no spatial structure on the latent space: an encoder $E$ maps a set of posed input images $\{I_i, \mathbf{r}_i\}_{i=1}^{N}$ (where $\mathbf{r}_i$ denotes the Pl\"ucker ray embedding of the $i$-th view) into a fixed-size set of 1D scene tokens $\mathbf{z} = E(\{I_i, \mathbf{r}_i\})$, and a decoder $D$ renders a novel view by conditioning on these tokens and a target ray embedding: $\hat{I} = D(\mathbf{z}, \mathbf{r}_t)$.

We observe that this paradigm opens up a fundamentally different design space for relighting.
Because the tokens encode an entire scene, including geometry and appearance, in a unified representation, relighting can be formulated as \emph{a single transformation on the tokens themselves}, rather than requiring explicit material decomposition, view-space generation, or re-rendering.
We propose a novel approach, \textbf{\shortname}, to exploit this property by introducing a Scene Token Editor $T$ that performs relighting as a direct transformation on the scene tokens: $\mathbf{z}' = T(\mathbf{z}, \boldsymbol{\ell})$, where $\boldsymbol{\ell}$ denotes the lighting signal encoded as a set of light-ray tokens (detailed in Sec.~\ref{subsec:architecture}).

A key design goal is that $T$ approximates a \emph{closed} operation on the latent token space $\mathcal{Z}$: if $\mathbf{z} \in \mathcal{Z}$, then $T(\mathbf{z}, \boldsymbol{\ell})$ should also lie in $\mathcal{Z}$ for any lighting signal $\boldsymbol{\ell}$, as the updated tokens still encode the same scene, just under a different lighting condition.
Ideally, the updated tokens $\mathbf{z}'$ are both decodable by $D$ into novel views and editable by further applications of $T$.
This makes the scene tokens a \emph{persistent, editable scene representation}: a user can apply an environment map, then add a point light, then adjust an area light, with each edit accumulating in token space and the decoder invoked only for the final rendering.
This stands in contrast to existing paradigms, where each lighting change requires either re-evaluating a rendering equation~\cite{TensoIR,Relightable3DGS}, re-running a full diffusion process~\cite{DiffusionRenderer,UniRelight,LightSwitch}, or, in hybrid methods such as RelitLRM~\cite{RelitLRM} and NeAR~\cite{NeAR}, regenerating appearance through a diffusion or generative branch even though the geometry is reusable.
To our knowledge, this idea has not been explored in prior work: existing latent scene representations have been used exclusively for view synthesis~\cite{SRT,LVSM,RayZer,SceneTok}, and relighting methods have not operated in such a token space.

Realizing this idea, however, poses several challenges.
The latent space is shaped by a reconstruction objective, and it is unclear whether coherent lighting transformations exist as learnable functions in this space.
Since geometry and appearance are entangled in the tokens, can the editor change illumination without corrupting the encoded scene structure?
And since the tokens carry no built-in 3D structure, can the model still support \emph{native 3D user interaction}, \eg, allowing a user to place a point light at a specific 3D location or orient an area light toward a surface, and respond correctly to such spatially grounded editing signals?
We address these challenges in Sec.~\ref{subsec:architecture}.

\subsection{\shortname~Architecture}
\label{subsec:architecture}

The full pipeline is illustrated in Fig.~\ref{fig:architecture}.
We adopt the encoder-decoder architecture of LVSM~\cite{LVSM} as our backbone and insert the Scene Token Editor $T$ between the encoder and decoder.
The editor design operates on generic scene tokens and is not tied to this specific choice.

\noindent\textbf{Encoder.}
The encoder $E$ is a transformer that maps a sparse set of $N$ posed input images into scene tokens $\mathbf{z} = \{\mathbf{z}_1, \dots, \mathbf{z}_K\} \in \mathbb{R}^{K \times d}$, where $K$ is the number of scene tokens and $d$ is the token dimension.
Each input image is patchified and concatenated with its Pl\"ucker ray embedding $\mathbf{r}_i$, forming a set of image tokens.
These image tokens are processed jointly with a fixed set of $K$ learnable empty tokens $\mathbf{z}_0$ through self-attention layers; the empty tokens aggregate spatial and photometric information from the input views and become the scene tokens $\mathbf{z} = E(\{I_i, \mathbf{r}_i\}_{i=1}^{N})$.

\noindent\textbf{Light signal tokenization.}
To address the challenge of interfacing 3D lighting specifications with a spatially unstructured representation (Sec.~\ref{subsec:latent_scene}), we encode all lighting signals into a unified token format.
We discretize the illumination into a collection of light rays sampled from the light source toward the visible scene area, producing a set of $L$ light tokens $\boldsymbol{\ell} = \{\boldsymbol{\ell}_1, \dots, \boldsymbol{\ell}_L\}$.
Each light token is parameterized by its spatial and emissive properties: $\boldsymbol{\ell}_i = (\mathbf{r}_i^{\ell}, \mathbf{c}_i, e_i)$, where $\mathbf{r}_i^{\ell}$ is the ray geometry in Pl\"ucker coordinates, $\mathbf{c}_i \in \mathbb{R}^3$ is the spectral intensity (color), and $e_i$ is the radiant power.
Following the Pl\"ucker ray convention, each light ray is represented as $\mathbf{r}_i^{\ell} = (\mathbf{d}_i, \mathbf{m}_i)$, where $\mathbf{d}_i$ is the unit direction and $\mathbf{m}_i = \mathbf{o}_i \times \mathbf{d}_i$ is the moment vector.

This representation naturally accommodates diverse light source types:
\begin{itemize}[itemsep=0pt,topsep=0pt,parsep=0pt,partopsep=0pt]
    \item \emph{Environment maps}: distant illumination with zero parallax with respect to the scene; we nullify the moment vector ($\mathbf{m}_i = \mathbf{0}$), reducing the Pl\"ucker representation to a pure directional encoding.
    \item \emph{Point lights}: rays originate from the source position and are directed toward the scene geometry area, so both the direction $\mathbf{d}_i$ and the moment $\mathbf{m}_i$ encode the light's 3D location.
    \item \emph{Area lights}: ray origins are sampled across the light's emitting surface, encoding both the position and spatial extent of the source.
\end{itemize}
This unified tokenization is what enables the native 3D user interaction discussed in Sec.~\ref{subsec:latent_scene}: regardless of whether the user specifies an environment map, places a point light at a particular 3D location, or orients an area light toward a surface, the lighting signal is converted into the same token format that the editor can process through a single attention mechanism.

\noindent\textbf{Scene Token Editor.}
The editor $T$ is a transformer that performs a light-conditioned transformation on the scene tokens.
Given scene tokens $\mathbf{z} = \{\mathbf{z}_1, \dots, \mathbf{z}_K\}$ and light tokens $\boldsymbol{\ell} = \{\boldsymbol{\ell}_1, \dots, \boldsymbol{\ell}_L\}$, the editor concatenates them into a single sequence and processes it through a stack of self-attention layers:
\begin{equation}
\mathbf{z}'_1, \dots, \mathbf{z}'_K, \boldsymbol{\ell}'_1, \dots, \boldsymbol{\ell}'_L = \text{Transformer}(\underbrace{\mathbf{z}_1, \dots, \mathbf{z}_K}_{\text{scene}},\; \underbrace{\boldsymbol{\ell}_1, \dots, \boldsymbol{\ell}_L}_{\text{light}}).
\label{eq:editor}
\end{equation}
The updated light tokens $\boldsymbol{\ell}'_1, \dots, \boldsymbol{\ell}'_L$ are discarded, and only the updated scene tokens $\mathbf{z}' = \{\mathbf{z}'_1, \dots, \mathbf{z}'_K\}$ are retained.
The use of self-attention, rather than cross-attention, is a deliberate design choice: it allows the light tokens to attend to the scene tokens, enabling the editor to learn where light interacts with scene geometry (\eg, where shadows should fall or highlights should appear), while simultaneously allowing scene tokens to attend to the light tokens to update their appearance accordingly.

\noindent\textbf{Decoder.}
The decoder $D$ is a ray-conditioned transformer that renders novel views from the scene tokens.
Given target camera Pl\"ucker ray embeddings $\mathbf{r}_t$, the decoder patchifies the target rays into query tokens and attends them to the scene tokens, producing a set of output tokens that are then mapped to pixels by a prediction head: $\hat{I} = D(\mathbf{z}', \mathbf{r}_t)$.
Rather than independently regressing each pixel patch from its corresponding output token with a per-token MLP readout as in LVSM~\cite{LVSM}, we attach a DPT head~\cite{DPT} to the decoder: it reassembles the decoder's output tokens into multi-scale feature maps and progressively fuses them through convolutional blocks into a dense, full-resolution image.
This dense-prediction head recovers noticeably sharper textures and high-frequency details than the per-token MLP readout.
Notably, the decoder is applied without modification to both the original scene tokens $\mathbf{z}$ (for novel view synthesis) and the edited tokens $\mathbf{z}'$ (for relighting).
The fact that the same decoder produces high-quality images in both cases suggests that the edited tokens remain approximately within the learned latent space, consistent with the design goal described in 
Sec.~\ref{subsec:latent_scene}.

\subsection{Progressive Relighting}
\label{subsec:progressive_relighting}

The Scene Token Editor $T$ (Sec.~\ref{subsec:latent_scene}) is designed so that its output remains in the same latent space as its input, enabling \emph{progressive relighting}: multiple lighting edits composed in token space before a single decoding pass.
Formally, given an initial scene representation $\mathbf{z} = E(\{I_i, \mathbf{r}_i\}_{i=1}^{N})$ and a sequence of $S$ lighting signals $\boldsymbol{\ell}^{(1)}, \dots, \boldsymbol{\ell}^{(S)}$, the editing process follows the recurrence
\begin{equation}
\mathbf{z}^{(1)} = T(\mathbf{z}, \boldsymbol{\ell}^{(1)}), \quad \mathbf{z}^{(s)} = T(\mathbf{z}^{(s-1)}, \boldsymbol{\ell}^{(s)}) \quad \text{for } s = 2, \dots, S,
\label{eq:progressive}
\end{equation}
and the final rendering is produced as $\hat{I} = D(\mathbf{z}^{(S)}, \mathbf{r}_t)$.
Because each intermediate $\mathbf{z}^{(s)}$ is intended to remain in $\mathcal{Z}$, it can be passed to both the decoder and the next editing step.
Each $\boldsymbol{\ell}^{(s)}$ can represent a different light source type: for instance, $\boldsymbol{\ell}^{(1)}$ may specify an environment map, $\boldsymbol{\ell}^{(2)}$ a key light placed at a particular 3D position, and $\boldsymbol{\ell}^{(3)}$ a fill light oriented toward the subject.
Fig.~\ref{fig:teaser} shows a qualitative example of this workflow.

This formulation supports \emph{progressive} lighting design: rather than specifying the complete target illumination upfront, the user builds it up one source at a time.
The decoder serves as a read-only preview: the next edit always starts from $\mathbf{z}^{(s)}$ itself, not from re-encoded pixels, avoiding the per-step reconstruction loss of a decode-and-re-encode pipeline.
To keep the edited tokens within the latent manifold over long chains, we train the editor to predict multiple sequential edits rather than a single transformation (Sec.~\ref{subsec:training}), which prevents the distributional drift that a single-step-trained editor would otherwise accumulate. As a result, token-space editing maintains higher fidelity than the pixel-space alternative throughout an interactive lighting session (Sec.~\ref{sec:experiments}).
To perform incremental editing, existing paradigms have to go through full recomputation at each step: inverse rendering must re-evaluate the rendering equation for the complete lighting configuration, diffusion-based methods must regenerate all views from scratch, and hybrid methods must re-run their appearance branch for each new condition.

\subsection{Training Objectives}
\label{subsec:training}

We optimize the encoder $E$ and the Scene Token Editor $T$ jointly, while keeping the pre-trained decoder $D$ frozen.
Fine-tuning $E$ is essential: the token space learned for novel view synthesis must adapt to support lighting transformations, and we find that freezing the encoder degrades relighting quality significantly (see Sec.~\ref{sec:experiments}).
The total training objective combines a photometric rendering loss $\mathcal{L}_{\text{render}}$ with an invariance regularization $\mathcal{L}_{\text{inv}}$:
\begin{equation}
\mathcal{L}_{\text{total}} = \mathcal{L}_{\text{render}} + \alpha \, \mathcal{L}_{\text{inv}},
\label{eq:total_loss}
\end{equation}
where $\alpha$ is a hyperparameter balancing the two terms.

\noindent\textbf{Rendering loss.}
To ensure the edited tokens $\mathbf{z}'$ correctly represent the scene under the target illumination, we supervise the decoded output against a ground-truth relit image $I_{\text{gt}}$.
The rendering loss combines a pixel-wise $\ell_2$ term with a perceptual term to capture high-frequency details:
\begin{equation}
\mathcal{L}_{\text{render}} = \| I_{\text{gt}} - \hat{I} \|_2^2 + \lambda \, \mathcal{L}_{\text{LPIPS}}(I_{\text{gt}}, \hat{I}),
\label{eq:render_loss}
\end{equation}
where $\hat{I} = D(\mathbf{z}', \mathbf{r}_t)$ is the image rendered by the decoder at target camera rays $\mathbf{r}_t$, $\mathcal{L}_{\text{LPIPS}}$ denotes the LPIPS perceptual similarity metric~\cite{LPIPS}, and $\lambda$ controls its relative weight.
To support progressive relighting (Sec.~\ref{subsec:progressive_relighting}), we do not restrict supervision to a single edit. Instead, we sample a random chain of lighting signals $\boldsymbol{\ell}^{(1)}, \dots, \boldsymbol{\ell}^{(S)}$, apply the editor recurrently in token space (Eq.~\ref{eq:progressive}), and decode the intermediate tokens $\mathbf{z}^{(s)}$ to apply the rendering loss at each step against the corresponding ground-truth relit image. Training the editor to predict multiple sequential edits keeps its outputs within the latent manifold over long chains and prevents the distributional drift that a single-step objective would accumulate.

\noindent\textbf{Lighting invariance loss.}
Since the encoder $E$ receives images captured under a particular source illumination, the scene tokens may entangle scene identity with source lighting effects.
To encourage the encoder to focus on the persistent scene content (geometry and materials) and leave all lighting-dependent variation to the editor $T$, we introduce a lighting invariance loss.
Given the same scene captured under two different lighting conditions $L_A$ and $L_B$, we penalize differences in the encoded tokens:
\begin{equation}
\mathcal{L}_{\text{inv}} = \| E(\{I_i^A, \mathbf{r}_i\}) - E(\{I_i^B, \mathbf{r}_i\}) \|_2^2,
\label{eq:inv_loss}
\end{equation}
where $\{I_i^A\}$ and $\{I_i^B\}$ are multi-view images of the same scene under lighting conditions $L_A$ and $L_B$, respectively.
This loss encourages the encoder to factor out the source illumination's contribution, so that the scene tokens $\mathbf{z}$ capture the persistent scene identity (geometry and materials) while leaving the editor responsible for injecting the target lighting.
We show in Sec.~\ref{sec:experiments} that this regularization yields a substantial improvement in relighting quality, and that it is as effective as an alternative design that explicitly supervises albedo prediction through an auxiliary decoder head.

\begin{figure}[t]
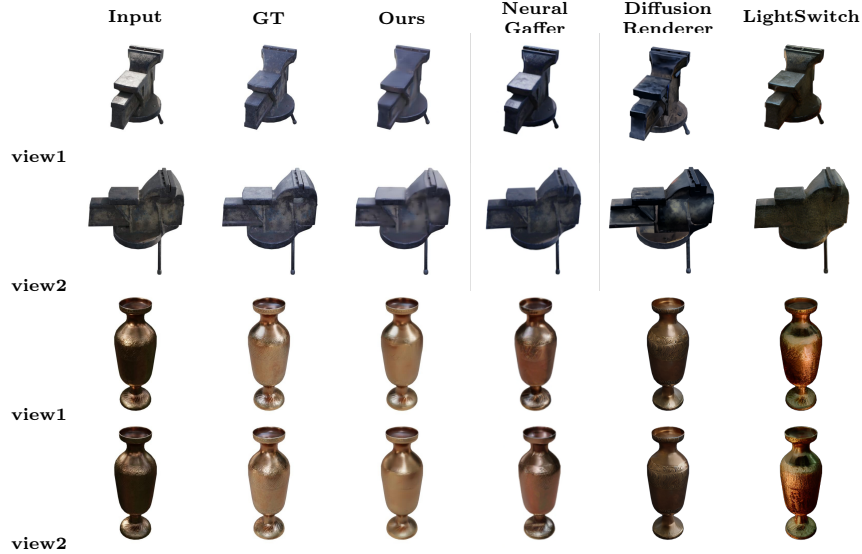

    \centering
    \setlength{\tabcolsep}{0.8pt}
    \renewcommand{\arraystretch}{0.0}
    \newcommand{\figfourimg}[1]{%
        \adjincludegraphics[trim={0} {0.05\height} {0} {0.05\height},clip,width=0.150\linewidth,height=0.150\linewidth]{figures_reOrganized/fig4/#1}%
    }
    \newcommand{\figfourimgII}[1]{%
        \adjincludegraphics[trim={0} {0.05\height} {0} {0.14\height},clip,width=0.150\linewidth,height=0.150\linewidth]{figures_reOrganized/fig4/#1}%
    }
    \newcommand{\colhead}[1]{\parbox[c][2.4\baselineskip][c]{0.150\linewidth}{\centering\scriptsize\textbf{\shortstack{#1}}}}
    \newcommand{\rowlab}[1]{{\scriptsize\textbf{#1}}}
    \footnotesize
    \resizebox{0.93\linewidth}{!}{%
    \begin{tabular}{lcccccc}
        & \colhead{Input} & \colhead{GT} & \colhead{Ours} & \colhead{Neural\\Gaffer} & \colhead{Diffusion\\Renderer} & \colhead{LightSwitch} \\[4pt]
        \rowlab{view1} & \figfourimg{r1c1} & \figfourimg{r1c2} & \figfourimg{r1c3} & \figfourimg{r1c4} & \figfourimg{r1c5} & \figfourimg{r1c6} \\
        \rowlab{view2} & \figfourimgII{r2c1} & \figfourimgII{r2c2} & \figfourimgII{r2c3} & \figfourimgII{r2c4} & \figfourimgII{r2c5} & \figfourimgII{r2c6} \\
        \rowlab{view1} & \figfourimg{r3c1} & \figfourimg{r3c2} & \figfourimg{r3c3} & \figfourimg{r3c4} & \figfourimg{r3c5} & \figfourimg{r3c6} \\
        \rowlab{view2} & \figfourimg{r4c1} & \figfourimg{r4c2} & \figfourimg{r4c3} & \figfourimg{split/r1c3.png} & \figfourimg{r4c5} & \figfourimg{r4c6} \\
    \end{tabular}%
    }
    \caption{\textbf{Qualitative comparison of multi-view relighting.} Columns correspond to methods and rows show two viewpoints for each of two scenes. Our method preserves appearance and lighting effects consistently across views, while baselines show artifacts such as residual source illumination and inconsistent highlights.}
    \label{fig:mv_qualitative}
\end{figure}

\section{Experiments}
\label{sec:experiments}






\subsection{Experiment Setup}
\label{sec:setup}

\begin{table}[t]
  \centering
  \caption{\textbf{Multi-view relighting accuracy.} ILR: image-level rescaling (per-image scale alignment). SLR: scene-level rescaling (single scale across all views per object). 
  }
  \label{tab:mv_relighting}
    \resizebox{0.85\linewidth}{!}{
  \begin{tabular}{lcccccc}
    \toprule
    \multirow{2}{*}{Method} & \multicolumn{3}{c}{Image-Level Rescaling (ILR)} & \multicolumn{3}{c}{Scene-Level Rescaling (SLR)} \\
    \cmidrule(lr){2-4} \cmidrule(lr){5-7}
     & PSNR$\uparrow$ & SSIM$\uparrow$ & LPIPS$\downarrow$ & PSNR$\uparrow$ & SSIM$\uparrow$ & LPIPS$\downarrow$ \\
    \midrule
    LightSwitch~\cite{LightSwitch}       & 21.22 & 0.868 & 0.105 & 21.19 & 0.868 & 0.131 \\
    Neural Gaffer~\cite{neuralGaffer}     & 28.40 & 0.947 & \textbf{0.030} & 28.23 & 0.955 & \textbf{0.037} \\
    DiffusionRenderer~\cite{DiffusionRenderer} & 26.39 & 0.951 & 0.038 & 26.24 & 0.923 & 0.086 \\
    Ours                                   & \textbf{30.48} & \textbf{0.954} & 0.045 & \textbf{30.36} & \textbf{0.917} & 0.086 \\
    \bottomrule
  \end{tabular}
  }
\end{table}

\noindent\textbf{Data.}
We train on a large-scale synthetic dataset of \textcolor{black}{20k} objects curated from Objaverse~\cite{Objaverse}, filtered to ensure sufficient geometric and material quality by \textcolor{black}{excluding objects with missing UV maps or non-PBR materials}. 
Each object is rendered in \textcolor{black}{Blender Cycles Engine} at $\textcolor{black}{512}{\times}\textcolor{black}{512}$ resolution from \textcolor{black}{80} viewpoints uniformly sampled on the upper hemisphere. 
For each object, we generate images under 16 illumination conditions: \textcolor{black}{4} HDR environment maps sampled from Polyhaven, \textcolor{black}{4} point light configurations, \textcolor{black}{2} area light configurations, and \textcolor{black}{2} combinations of multiple light sources, yielding approximately \textcolor{black}{20} million images in total. 
We evaluate on a held-out set of \textcolor{black}{900} objects (\textcolor{black}{700} from Objaverse and \textcolor{black}{200} from the Polyhaven model library), with no overlap with the training set. Each test object is rendered under \textcolor{black}{8} lighting conditions unseen during training. 

\noindent\textbf{Evaluation Protocol.}
We assess two tasks:
(1)~\textit{Multi-view relighting}, where the model relights all input views at their original poses under a new target illumination, isolating relighting quality from view synthesis;
and (2)~\textit{Novel-view relighting}, where the model renders relit images from viewpoints unseen during encoding, evaluating joint relighting and view synthesis.
For multi-view relighting, the model receives \textcolor{black}{4} input views. 
For novel-view relighting, the model receives \textcolor{black}{8} sparse input views and is evaluated on \textcolor{black}{8} held-out viewpoints. 
Following LightSwitch~\cite{LightSwitch}, we account for the inherent albedo-exposure ambiguity by applying a least-squares scale alignment to the prediction before computing PSNR, SSIM, and LPIPS~\cite{LPIPS}.
For multi-view relighting, we report metrics under two rescaling protocols: image-level rescaling (ILR), which computes an optimal scale per image, and scene-level rescaling (SLR), which computes a single scale across all views of an object. 

\noindent\textbf{Baselines.}
For multi-view relighting, we compare against diffusion-based methods that can relight individual views: Neural Gaffer~\cite{neuralGaffer}, DiffusionRenderer~\cite{DiffusionRenderer}, and LightSwitch~\cite{LightSwitch}.
For novel-view relighting, we compare against inverse rendering methods that perform per-scene optimization: TensoIR~\cite{TensoIR} and NVDiffrecMC~\cite{NVDiffrecMC}; 
as well as the multi-view diffusion method LightSwitch~\cite{LightSwitch}.
We did not compare against RelitLRM~\cite{RelitLRM}, whose code and pretrained models remain unavailable, or NeAR~\cite{NeAR}, whose implementation was released after our experimental evaluation was completed.

\noindent\textbf{Implementation Details.} Our backbone follows the encoder-decoder architecture of LVSM~\cite{LVSM}, with \textcolor{black}{12} transformer layers each in the encoder, decoder, and Scene Token Editor (token dimension $d{=}\textcolor{black}{768}$), representing each scene with $K{=}\textcolor{black}{3072}$ tokens. For the decoder readout, we replace the per-token MLP of LVSM with a DPT head~\cite{DPT} to recover sharper textures and high-frequency details. We provide the full set of training hyperparameters and schedule in the supplementary material.

\subsection{Results}

\begin{table}[t]
  \centering
  \caption{\textbf{Novel-view relighting accuracy.}
    \#Input denotes how many source views each method uses;
    our method requires significantly fewer.
    }
  \label{tab:nv_relighting}
  \resizebox{\linewidth}{!}{%
  \begin{tabular}{lc ccc ccc ccc}
    \toprule
    \multirow{2}{*}{Method}
      & \multirow{2}{*}{\#Input}
      & \multicolumn{3}{c}{Synthetic}
      & \multicolumn{3}{c}{Objects-with-Lighting}
      & \multicolumn{3}{c}{Stanford-ORB} \\
    \cmidrule(lr){3-5} \cmidrule(lr){6-8} \cmidrule(lr){9-11}
      &
      & PSNR$\uparrow$ & SSIM$\uparrow$ & LPIPS$\downarrow$
      & PSNR$\uparrow$ & SSIM$\uparrow$ & LPIPS$\downarrow$
      & PSNR$\uparrow$ & SSIM$\uparrow$ & LPIPS$\downarrow$ \\
    \midrule
    LightSwitch~\cite{LightSwitch}
      & 16 & 21.61 & 0.86 & 0.13
             & 25.43 & 0.84  & 0.30
             & \textbf{32.02} & \textbf{0.98} & 0.03 \\
    NVDiffrecMC~\cite{NVDiffrecMC}
      & 50 & 22.95 & 0.86 & 0.10
             & 20.24 & 0.73  & 0.40
             & 31.60 & 0.97  & 0.04 \\
    TensoIR~\cite{TensoIR}
      & 50 & 26.17 & \textbf{0.92} & 0.07
             & 26.12 & 0.77  & 0.38
             & 25.27     & 0.94     & 0.06 \\
    Ours
      &  8 & \textbf{27.76} & 0.91 & \textbf{0.06}
             & \textbf{26.76} & \textbf{0.92} & 0.36
             & 31.01 & 0.97  & 0.03 \\
    \bottomrule
  \end{tabular}%
  }
\end{table}

\begin{figure}[t]
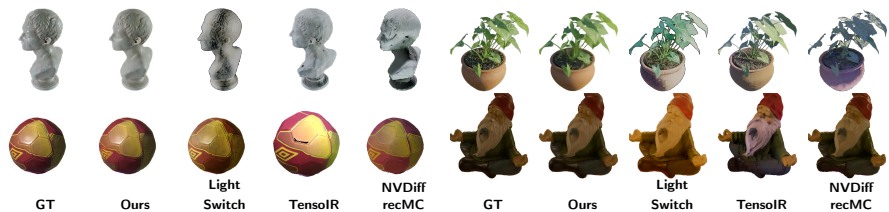

  \setlength{\tabcolsep}{0.6pt}%
  \renewcommand{\arraystretch}{0.0}%
    \newcommand{\figfiveimg}[1]{%
    \adjincludegraphics[trim={0.10\width} {0.10\height} {0.10\width} {0.10\height},clip,width=0.0935\linewidth,height=0.0935\linewidth]{figures_reOrganized/fig5/#1}%
  }%
  \newcommand{\figfiveimgB}[1]{%
    \adjincludegraphics[trim={0.20\width} {0.20\height} {0.20\width} {0.20\height},clip,width=0.0935\linewidth,height=0.0935\linewidth]{figures_reOrganized/fig5_additional/#1}%
  }%
  \newcommand{\figfiveimgC}[1]{%
    \adjincludegraphics[trim={0.15\width} {0.15\height} {0.15\width} {0.15\height},clip,width=0.0935\linewidth,height=0.0935\linewidth]{figures_reOrganized/fig5_additional/#1}%
  }%
  \newcommand{\figfiveblank}{\makebox[0.0935\linewidth][c]{\rule{0pt}{0.0935\linewidth}}}%
  \newcommand{\figfivelab}[1]{\parbox[t]{0.0935\linewidth}{\centering\tiny\sffamily\bfseries\shortstack{#1}}}%
  \begin{tabular}{ccccc ccccc}
    \figfiveimg{r2c1} & \figfiveimg{r2c2}  & \figfiveimg{r2c3} & \figfiveimg{r2c4} & \figfiveimg{r2c5} &
    \figfiveimg{r4c1} & \figfiveimg{r4c2}  & \figfiveimg{r4c3} & \figfiveimg{r4c4} & \figfiveimg{r4c5} \\
    \figfiveimgB{sto_r1c1} & \figfiveimgB{sto_r1c2} & \figfiveimgB{sto_r1c3} & \figfiveimgB{sto_r1c4} & \figfiveimgB{sto_r1c5} &
    \figfiveimgC{sto_r3c1} & \figfiveimgC{sto_r3c2} & \figfiveimgC{sto_r3c3} & \figfiveimgC{sto_r3c4} & \figfiveimgC{sto_r3c5} \\
    \figfivelab{GT} & \figfivelab{Ours} & \figfivelab{Light\\Switch} & \figfivelab{TensoIR} & \figfivelab{NVDiff\\recMC} &
    \figfivelab{GT} & \figfivelab{Ours} & \figfivelab{Light\\Switch} &\figfivelab{TensoIR} & \figfivelab{NVDiff\\recMC} \\

  \end{tabular}
  \captionof{figure}{\textbf{Visual comparison of novel-view relighting.}}
  \label{fig:nvs_relight}
\end{figure}

\noindent\textbf{Multi-view Relighting.} As shown in Tab.~\ref{tab:mv_relighting} and Fig.~\ref{fig:mv_qualitative}, our method achieves the highest PSNR and SSIM under both rescaling protocols, exceeding Neural Gaffer by \textcolor{red}{2.1}~dB in PSNR and outperforming all baselines in SSIM (0.954).
Our SSIM and LPIPS are identical under ILR and SLR, and PSNR drops by only 0.12~dB, indicating consistent relighting across views.
On LPIPS, our model (0.045) is slightly outperformed by Neural Gaffer (0.030), the best-performing method on this metric.
A possible explanation is that as a diffusion-based model, Neural Gaffer benefits from a generative prior that produces perceptually sharp textures, which LPIPS rewards. 
Notably, DiffusionRenderer achieves a comparable ILR LPIPS (0.038) but degrades substantially under SLR (0.086), while our LPIPS remains unchanged (0.045 under both protocols), reflecting more 
consistent perceptual quality across views.
Because our model processes all input views jointly through shared scene tokens, it can leverage aggregated multi-view information to separate the object's intrinsic appearance from source lighting effects such as specular highlights and cast shadows. In contrast, single-image baselines (NeuralGaffer) relight each view independently, which limits their ability to resolve such ambiguities.
\noindent\textbf{Novel-view Relighting.}
\label{sec:nv_relighting}
As shown in Tab.~\ref{tab:nv_relighting} and Fig.~\ref{fig:nvs_relight}, our method achieves the highest PSNR (27.76~dB) and best LPIPS (0.069) among synthetic data, and \textcolor{black}{competitive score in realworld bench marks}, while requiring only 8 input views, while other baselines require 30-50 views as input.
On SSIM, TensoIR achieves the highest score for synthetic data(0.928 vs.\ ours 0.917). This is consistent with the nature of inverse rendering: by explicitly recovering geometry and materials through dense-view optimization, TensoIR can produce structurally precise reconstructions that SSIM rewards. Our method trades this marginal structural advantage for a large reduction in input requirements and the elimination of per-scene optimization.
\subsection{Ablation Studies}
\label{sec:ablations}

\noindent\textbf{Model Design.}
We ablate two key design choices in Tab.~\ref{tab:ablation}: whether to finetune the encoder, and the form of regularization that encourages the encoder to separate scene identity from source lighting. Freezing the encoder, as done
\begin{wraptable}{r}{0.465\linewidth} 
\vspace{-25pt}
    \centering
  \captionof{table}{Ablation study on encoder training and lighting invariance regularization for novel-view relighting.}
  \label{tab:ablation}
  \resizebox{\linewidth}{!}{%
  \begin{tabular}{llccc}
    \toprule
    Encoder & Regularization & PSNR$\uparrow$ & SSIM$\uparrow$ & LPIPS$\downarrow$ \\
    \midrule
    Frozen   & None                          & 25.06 & 0.910 & 0.073 \\
    Unfrozen & None                          & 26.12 & 0.927 & 0.046 \\
    Unfrozen & Explicit (predict albedo)     & \textbf{27.09} & 0.934 & 0.057 \\
    Unfrozen & Implicit (invariance loss)    & 27.05 & \textbf{0.952} & \textbf{0.045} \\
    \bottomrule
  \end{tabular}%
  }
  \vspace{-16pt}
\end{wraptable}
in LVSM for novel view synthesis, leads to a 1.1~dB PSNR drop compared to the unfrozen baseline (25.06 vs.\ 26.12). This confirms that the token space learned for view synthesis alone does not naturally support lighting transformations; the encoder must adapt to produce tokens that are amenable to editing.
Adding regularization yields a further improvement of approximately 1~dB. Two forms achieve comparable PSNR: explicit albedo prediction through an auxiliary decoder head (27.09) and our proposed implicit invariance loss (27.05). However, the invariance loss achieves notably better SSIM (0.952 vs.\ 0.934) and LPIPS (0.045 vs.\ 0.057), while requiring no ground-truth albedo annotations. We adopt the invariance loss as our default.


\noindent\textbf{Progressive Relighting.}
A key property of our framework is that the Scene Token Editor's output remains in the same latent space as its input (Sec.~\ref{subsec:progressive_relighting}), enabling multiple lighting edits to be chained in token space without decoding to pixels between steps.
To make the editor robust to such chaining, we 
train it to predict multiple sequential edits rather than a single transformation: at training time we apply a random chain of lighting edits directly in token space and supervise the decoded output at each step. This forces the editor to keep its outputs within the latent manifold over long editing chains, preventing the distributional drift that a single-step-trained editor would otherwise accumulate.
We evaluate 
the model's ability to preserve quality under repeated edits using the following protocol. For each test object, we randomly sample a pool of 5 lighting conditions and render the corresponding ground-truth relit images with the Blender Cycles engine. At each editing step, one lighting condition is randomly selected from this pool as the new target, and the image rendered under the selected condition serves as the ground truth for that step. 
\begin{wrapfigure}{r}{0.469\linewidth}
\vspace{-8pt}
  \centering
  \includegraphics[width=\linewidth]{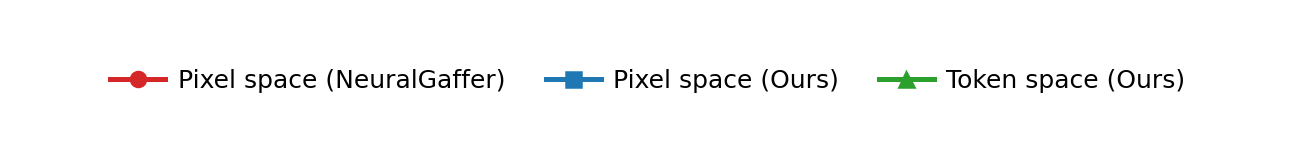}\par
  \includegraphics[width=\linewidth]{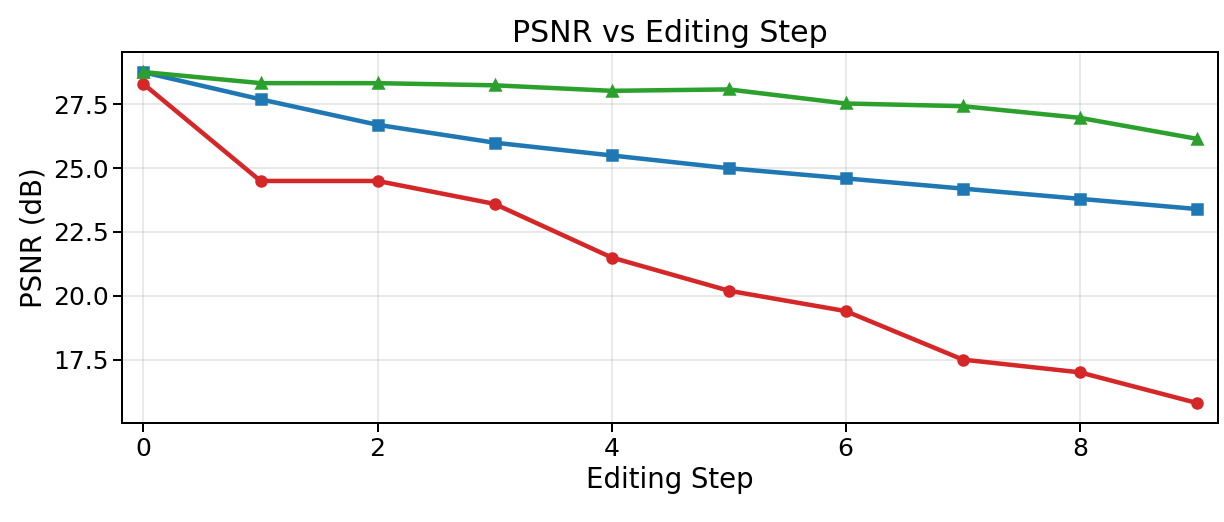}\par
  \includegraphics[width=\linewidth]{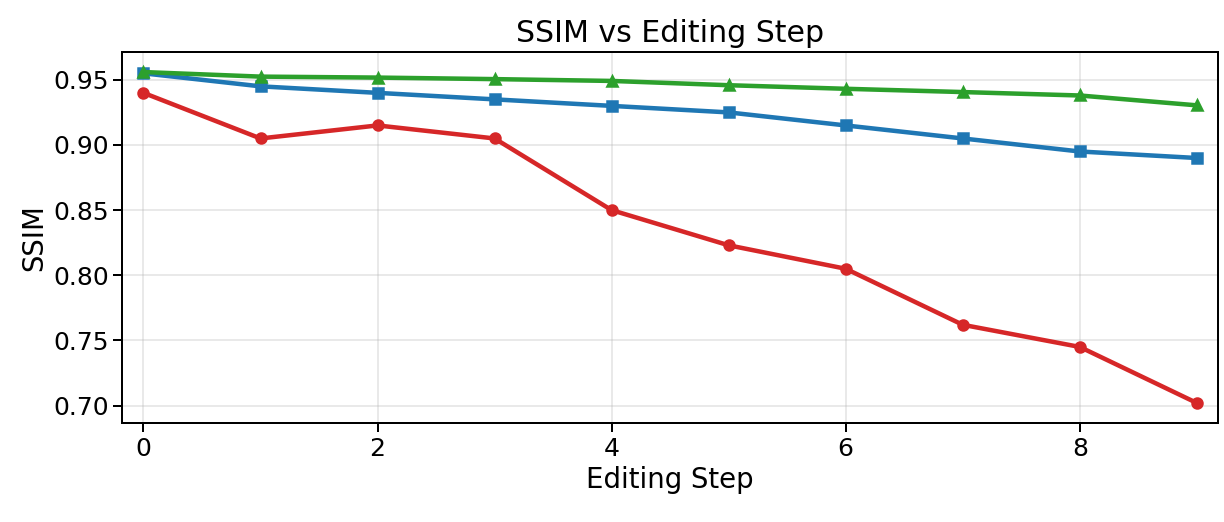}\par
  \setlength{\tabcolsep}{0.3pt}%
  \renewcommand{\arraystretch}{0.0}%
  \newcommand{\fsixcell}[1]{\includegraphics[width=1cm,height=1cm]{figures_reOrganized/fig6/interaction/#1}}%
  \newcommand{\fsixrow}[1]{\resizebox{\linewidth}{!}{\begin{tabular}{cccccccccc}#1\end{tabular}}\par}%
  \newcommand{\fsixlab}[1]{{\tiny\sffamily\bfseries #1}\par}%
  \fsixlab{Ours (token-space)}%
  \fsixrow{\fsixcell{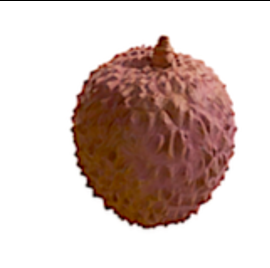} & \fsixcell{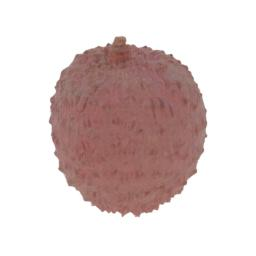} & \fsixcell{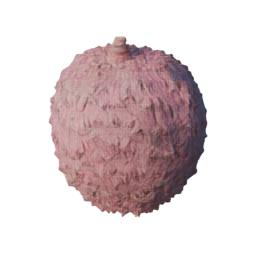} & \fsixcell{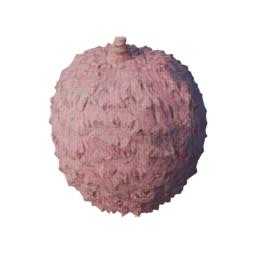} & \fsixcell{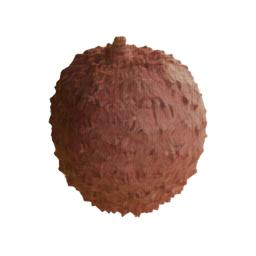} & \fsixcell{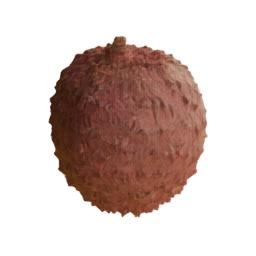} & \fsixcell{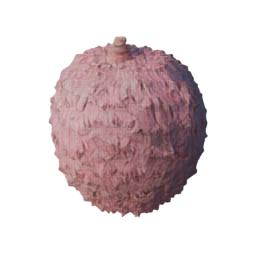} & \fsixcell{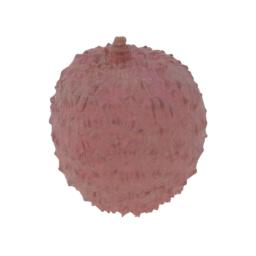} & \fsixcell{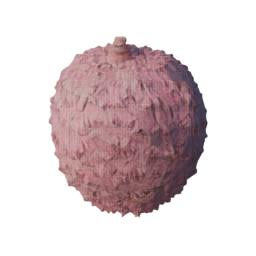} & \fsixcell{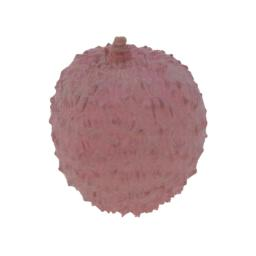}}%
  \fsixlab{Ours (pixel-space)}%
  \fsixrow{\fsixcell{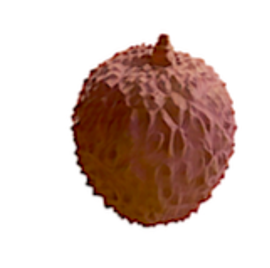} & \fsixcell{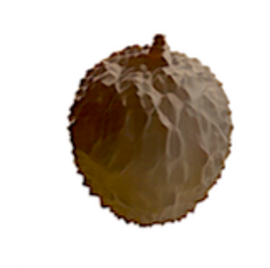} & \fsixcell{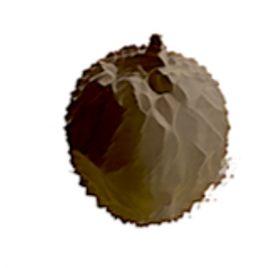} & \fsixcell{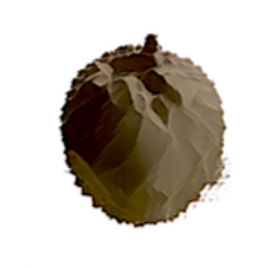} & \fsixcell{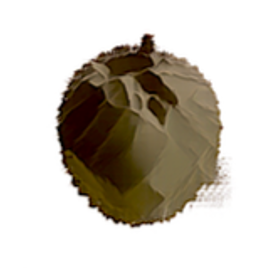} & \fsixcell{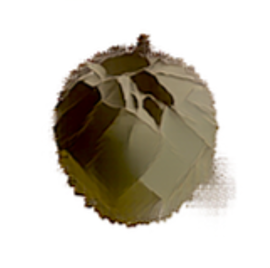} & \fsixcell{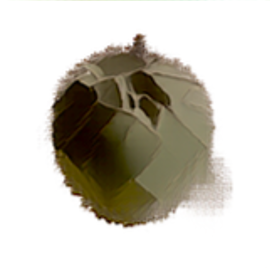} & \fsixcell{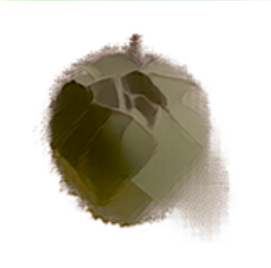} & \fsixcell{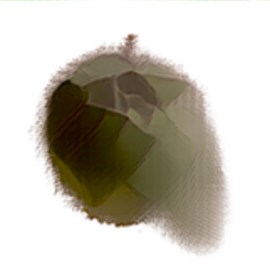} & \fsixcell{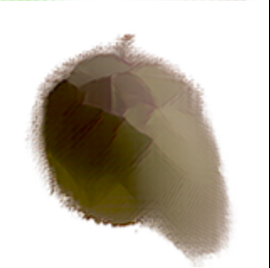}}%
  \fsixlab{NeuralGaffer (pixel-space)}%
  \fsixrow{\fsixcell{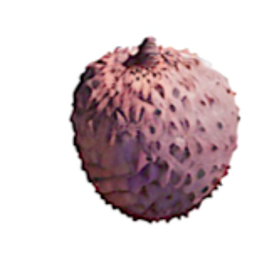} & \fsixcell{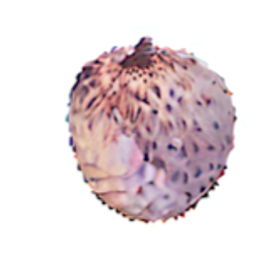} & \fsixcell{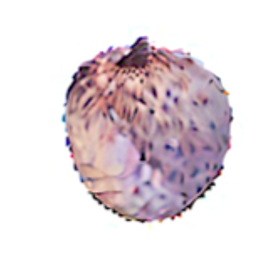} & \fsixcell{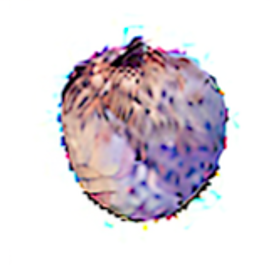} & \fsixcell{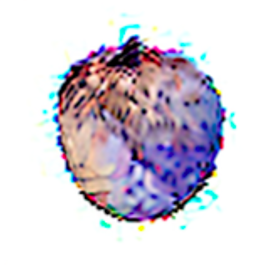} & \fsixcell{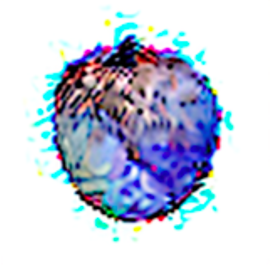} & \fsixcell{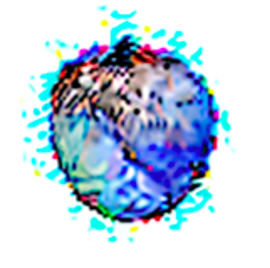} & \fsixcell{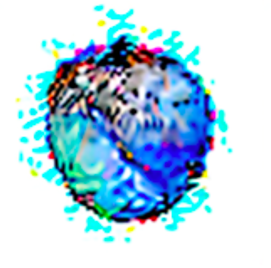} & \fsixcell{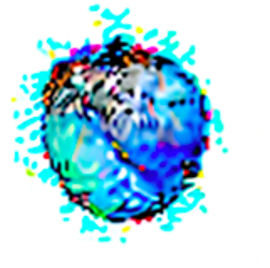} & \fsixcell{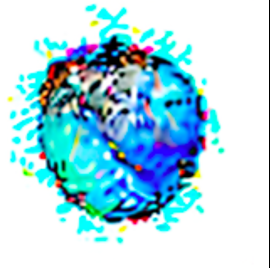}}%
  \captionof{figure}{Progressive relighting ablation. \textit{Top plots:} PSNR and SSIM over multiple editing steps, comparing token-space editing (ours) with pixel-space decode-and-re-encode. Token-space editing preserves fidelity over early steps, while pixel-space editing degrades from the first step onward. \textit{Bottom:} visual comparison across editing steps showing the same trend.}
  \label{fig:progressive_editing}
  \vspace{-70pt}
\end{wrapfigure}
We compare two editing strategies over 10 sequential relighting steps:
(1)~\emph{token-space editing}, where the scene tokens from the previous step are passed directly to the editor together with the new lighting signal; and 
(2)~\emph{pixel-space editing}, where the decoded image from the previous step is re-encoded by the encoder into scene tokens before the editor is applied with the new lighting signal.
As shown in Fig.~\ref{fig:progressive_editing}, with this multi-step training, token-space editing is consistently better than pixel-space editing across all 10 steps: it preserves fidelity over long chains, whereas pixel-space editing accumulates reconstruction drift from the very first step due to the lossy decode-and-re-encode round trip. This confirms that, throughout the practical range of interactive lighting design, chaining edits directly in token space is the preferred strategy.

\section{Conclusion}
\label{sec:conclusion}

We have presented LumiTokens, a framework that formulates 3D relighting as a direct transformation in a learned latent token space.
By introducing the Scene Token Editor, which processes unstructured scene tokens jointly with Pl\"ucker-parameterized light-ray tokens through self-attention, our method supports environment maps, point lights, and area lights through a unified interface and enables progressive, composable lighting edits entirely in token space. Experiments show that LumiTokens achieves competitive or superior relighting quality compared to inverse rendering and diffusion-based methods that require significantly more input views and costly per-scene optimization. Extending the framework to real-world captures with complex materials and imperfect poses, and training the editor on multi-step chains to improve robustness over longer editing sequences, are promising directions for future work.

%
%
\bibliographystyle{splncs04}
\bibliography{main}

\clearpage
\appendix
\section*{Supplementary Material}
\section{Additional Implementation Details}
\label{sec:supp:implementation}


\subsection{Dataset Creation}

\textbf{Single Object Data:}
To train LumiTokens, we select 20K different objects from objaverse, curated by neuralGaffer \cite{neuralGaffer}. We render the objects under several different lighting variations, following the strategy described by DilightNet \cite{dilightNet}.
Specifically, we render the objects under the following lighting variations:

\textit{Point light sources:} We sample 1 point light source randomly positioned on the upper hemisphere of the object. It is distanced [3.0, 5.0] from the object, then intensity uniformly sampled from [500,1500].

\textit{Area light sources:} We sample 1 area light source on the upper hemisphere of the object. It is distanced [3.0, 5.0] from the object, then intensity uniformly sampled from [500,1500]. The area light is square-shaped with side length [0.5, 4.0] and oriented towards the object.

\textit{Environment map:} We sample 1 environment map from the open-source HDRIs collection provided by PolyHaven. The environment map is randomly rotated around the object to provide more diversity.

\textit{Combined lights:} We create samples to simulate complex lighting conditions by combining multiple light sources. We sample 1 environment map, up to 3 different pointlight sources, and up to 1 area light source, and render the object under the combined illumination.

For each object we sample 4 different evironment map, 4 different point light configurations, and 1 area light configureation, as well as 4 combined light configurations.

We sample 50 frames of camera poses along a randomly sampled trajectory circling the object on the upper hemisphere, and another 30 frames uniformly distributed on the upper hemisphere for rendering.

This in total gives us 20,800,000 image samples for training.

\noindent\textbf{Scene Level Data:} 
To further tune the model on scene-level data, which includes complex lighting effects such as inter-reflections as well as hard and soft shadows, we generate 3,000 scenes composed of multiple objects placed on a ground plane.


We follow the scene construction strategy introduced by GenLit \cite{GenLit}. Specifically, we curate a list of 2,000 high-quality object models from the Objaverse-LVIS split, together with 200 high-quality 3D models from PolyHaven's public model collection. For each scene, we randomly sample one object from PolyHaven due to its high-quality textures and materials, and up to seven additional objects from the Objaverse-LVIS split to increase object diversity. The objects are randomly scaled within a range of [0.5, 1.5] in world units and randomly rotated to arbitrary orientations.

Next, we create a planar ground surface and apply a texture randomly sampled from the PolyHaven texture collection. The objects are then placed on the ground plane within the spatial range of [-3.0, 3.0] along the $x$-$y$ axes of the world coordinate system. Object positions are sampled with increased importance sampling around the center of the ground plane to encourage spatial diversity. If a newly placed object collides with an existing object, its position is resampled until a collision-free placement is obtained.

\subsection{Light Signal Tokenization}

\begin{figure}[t]
    \centering
    \includegraphics[width=1.0\linewidth]{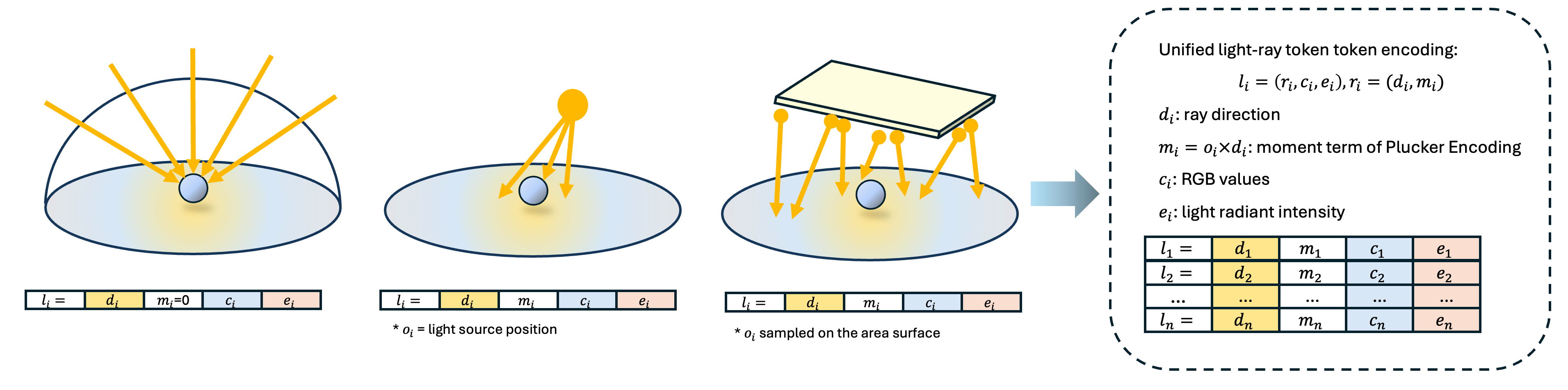}
    \caption{Light tokenization: different light sources (environment map, point light, area light) are converted into a unified ray-token representation.}
    \label{fig:supp:light_tokenization}
\end{figure}

To interface arbitrary lighting specifications with the Scene Token Editor, we represent illumination as a collection of light rays. As illustrated in Figure \ref{fig:supp:light_tokenization},
each light ray is encoded as a token
\begin{equation}
\boldsymbol{\ell}_i = (\mathbf{r}_i^{\ell}, \mathbf{c}_i, e_i),
\end{equation}
where $\mathbf{r}_i^{\ell}$ denotes the ray geometry expressed in Pl\"ucker coordinates, $\mathbf{c}_i$ is the RGB spectral intensity, and $e_i$ represents the radiant power. The ray geometry is parameterized as
\begin{equation}
\mathbf{r}_i^{\ell} = (\mathbf{d}_i, \mathbf{m}_i),
\end{equation}
where $\mathbf{d}_i$ is the unit direction and $\mathbf{m}_i = \mathbf{o}_i \times \mathbf{d}_i$ is the moment vector determined by the ray origin $\mathbf{o}_i$.

\begin{itemize}[itemsep=0pt,topsep=0pt,parsep=0pt,partopsep=0pt]
    \item \emph{Environment map.} 
    As shown on Figure \ref{fig:supp:light_tokenization}, partA, an environment map is represented as a batch of directional rays surrounding the scene with $\mathbf{m}_i = \mathbf{0}$, corresponding to illumination from infinity.
    
    \item \emph{Point light.}
    As shown on Figure \ref{fig:supp:light_tokenization}, partB, a point light is represented as a ray with the origin $\mathbf{o}_i$ at the light position and the direction $\mathbf{d}_i$ sampled from the origin to anywhere in the scene.

    \item  \emph{Area light.}
    As shown on Figure \ref{fig:supp:light_tokenization}, partC, an area light is represented as a batch of rays with the origin $\mathbf{o}_i$ sampled from the area light surface and the direction $\mathbf{d}_i$ sampled from the origin to anywhere in the scene.
\end{itemize}

This representation naturally unifies multiple lighting types. Environment maps are represented as directional rays with $\mathbf{m}_i = \mathbf{0}$, corresponding to illumination from infinity. Point lights emit rays from a shared origin corresponding to the light position, such that both the direction $\mathbf{d}_i$ and moment $\mathbf{m}_i$ encode the spatial location of the light source. Area lights are represented by sampling ray origins $\mathbf{o}_i$ across the emitting surface, allowing the representation to capture both the spatial extent and orientation of the light.

By converting all lighting signals into a common tokenized ray representation, the Scene Token Editor can process lighting and scene information jointly through self-attention.

\subsection{Model Training}
We begin by training the reconstruction backbone on our rendered multi-view object dataset for 50k steps. This initial phase is conducted without varying lighting conditions to ensure a stable initialization that aligns with our specific camera intrinsic and extrinsic distributions.

Subsequently, we jointly train the Scene Token Editor and the encoder on the multi-view object data for 100k steps, incorporating diverse lighting conditions. To enhance the model's ability to capture the complex illumination effects required for scene-level relighting, we conclude with a fine-tuning stage with scene-level data added into the training set.

All stages are trained on \textcolor{black}{8} NVIDIA \textcolor{black}{H100 80GB} GPUs with a batch size of \textcolor{black}{128} using the AdamW optimizer with a learning rate of \textcolor{black}{1e-4}. The loss weights are set to $\lambda{=}\textcolor{black}{0.9}$ for the perceptual term and $\alpha{=}\textcolor{black}{0.1}$ for the invariance regularization.
\section{Additional Evaluations}
\label{sec:supp:additional_evaluations}

\subsection{Runtime Costs}

\begin{table}[t]
  \centering
  \caption{Runtime and memory comparison. End-to-end inference and relight-only times (sec), peak GPU memory (MB), and CPU Memory (MB). Evaluated on a NVIDIA A100 GPU.
  }
  \label{tab:supp:runtime}
  \setlength{\tabcolsep}{3.5pt}
  \begin{tabular}{lccccc}
    \toprule
    Method & End-to-end & Relight Only & Peak GPU Mem. & CPU Memory \\
    \midrule
    \rowcolor{blue!15}
    Ours & \textbf{3.79} & \textbf{3.77} & \textbf{6108} & 3015 \\
    NeuralGaffer & 16.85 & 16.85 & 24064 & 5622 \\
    DiffusionRenderer & 22.81 & 14.36 & 28755 & 7460 \\
    LightSwitch & 532.33 & 127.10 & 22696 & 2882 \\
    TensoIR & 3952.91 & 294.40 & 10979 & 5038 \\
    NVDiffrecMC & 1691.75 & 225.41 & 6312 & \textbf{2258} \\
    \bottomrule
  \end{tabular}
\end{table}

We provide additional runtime cost of our model against other methods in Tab.~\ref{tab:supp:runtime}. We evaluate the inference time of relighting 100 novel views of a single object under a new lighting condition on an A100 GPU, in addition to GPU memory consumption.

End-to-end inference time refers to the inference time it takes to finish reconstructing (if such stage exists) the object and then relight 100 novel views. Relight only inference time refers to the inference time it takes to relight a scene, if a scene and its property is already reconstructed and optimized, and only lighting needs to be changed.

The result shows that our model is significantly faster than other methods, and the memory consumption is also comparable to other methods.



\section{Additional Qualitative Results}
\label{sec:supp:additional_qualitative}

\begin{figure}[t]
    \centering
    \begin{minipage}{0.98\linewidth}
        \centering
        \includegraphics[width=\linewidth]{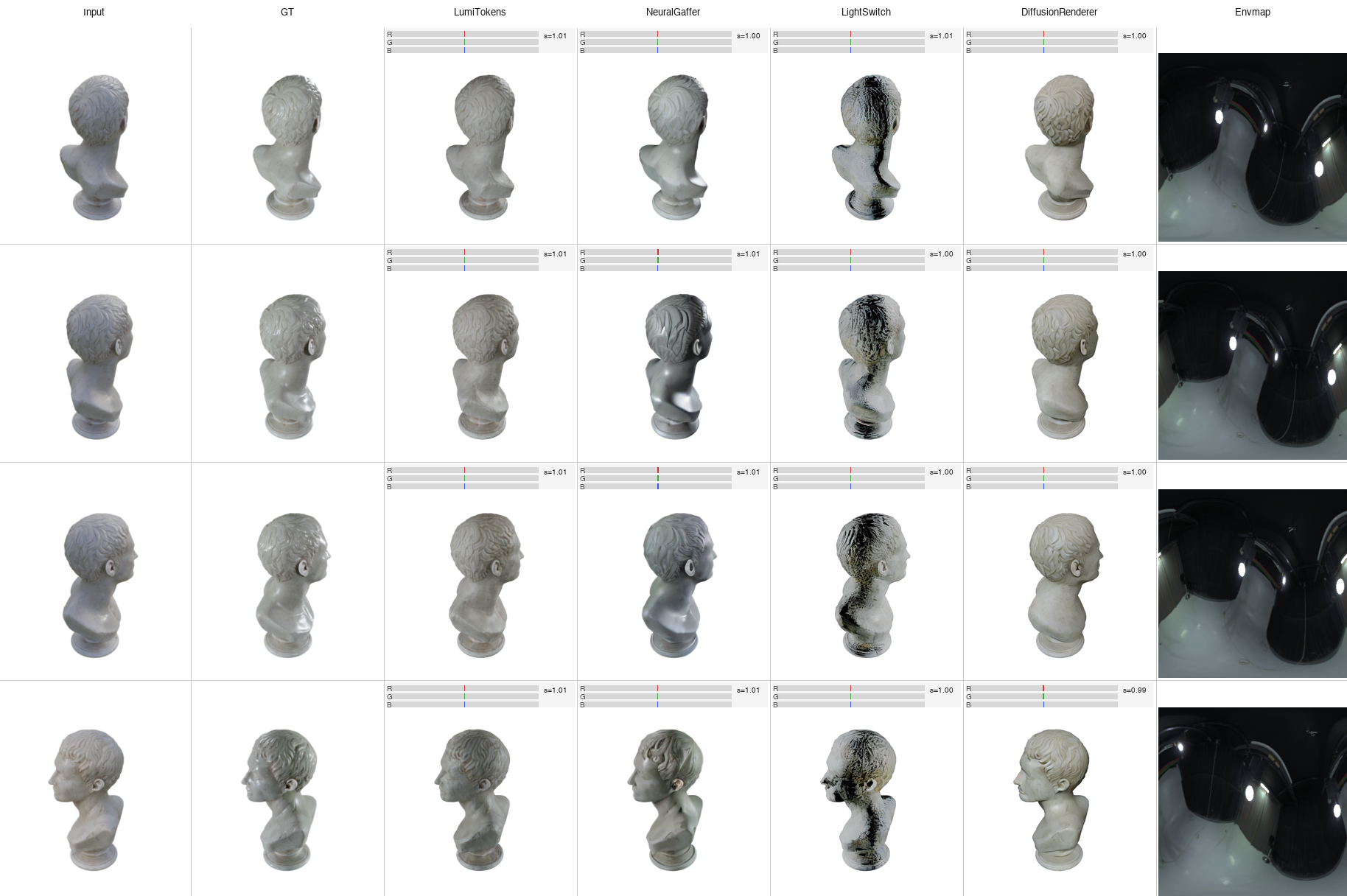}
        \subcaption{Marble bust.}
    \end{minipage}\hfill
    \begin{minipage}{0.98\linewidth}
        \centering
        \includegraphics[width=\linewidth]{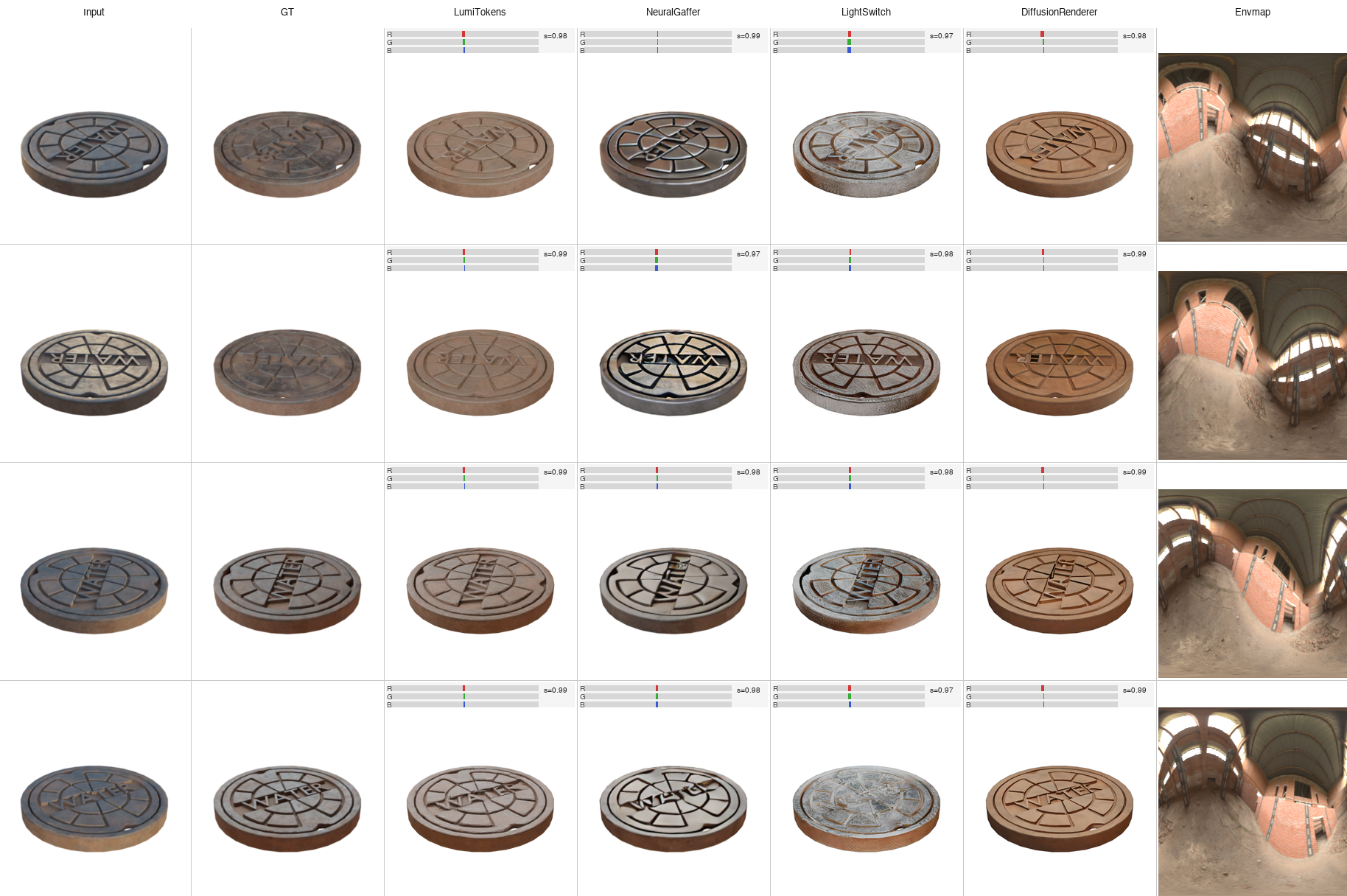}
        \subcaption{Water manhole cover.}
    \end{minipage}
    \caption{Multi-view relighting comparisons. Per-view optimal tone scales (shown as sliders above each image) reveal that LumiToken exhibits more stable scaling across views than baselines, demonstrating better cross-view consistency.}
    \label{fig:supp:mv_relight_comparisons_part1}
\end{figure}

\begin{figure}
    \centering
    \medskip
    \begin{minipage}{0.98\linewidth}
        \centering
        \includegraphics[width=\linewidth]{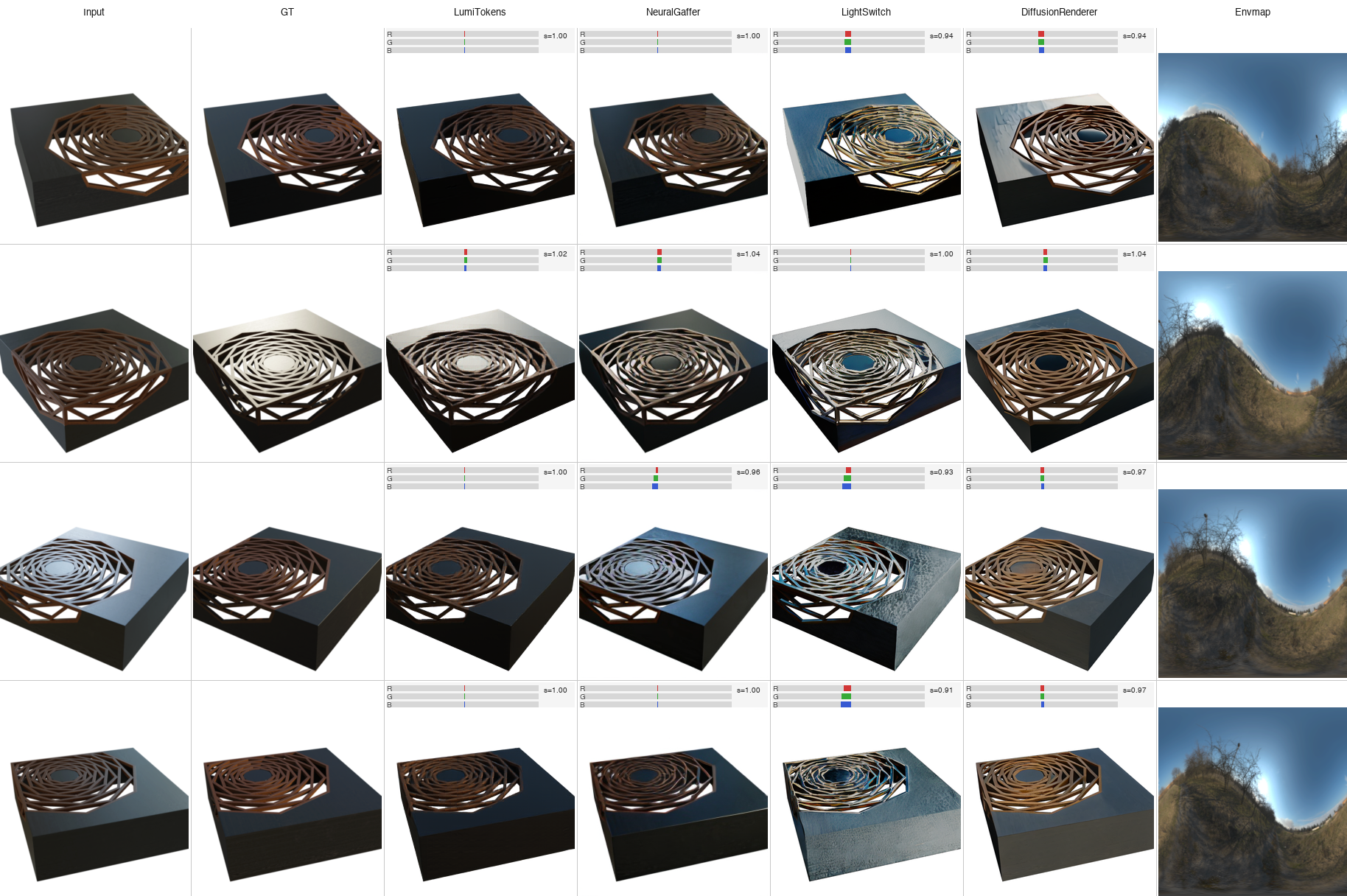}
        \subcaption{Modern coffee table.}
    \end{minipage}\hfill
    \begin{minipage}{0.98\linewidth}
        \centering
        \includegraphics[width=\linewidth]{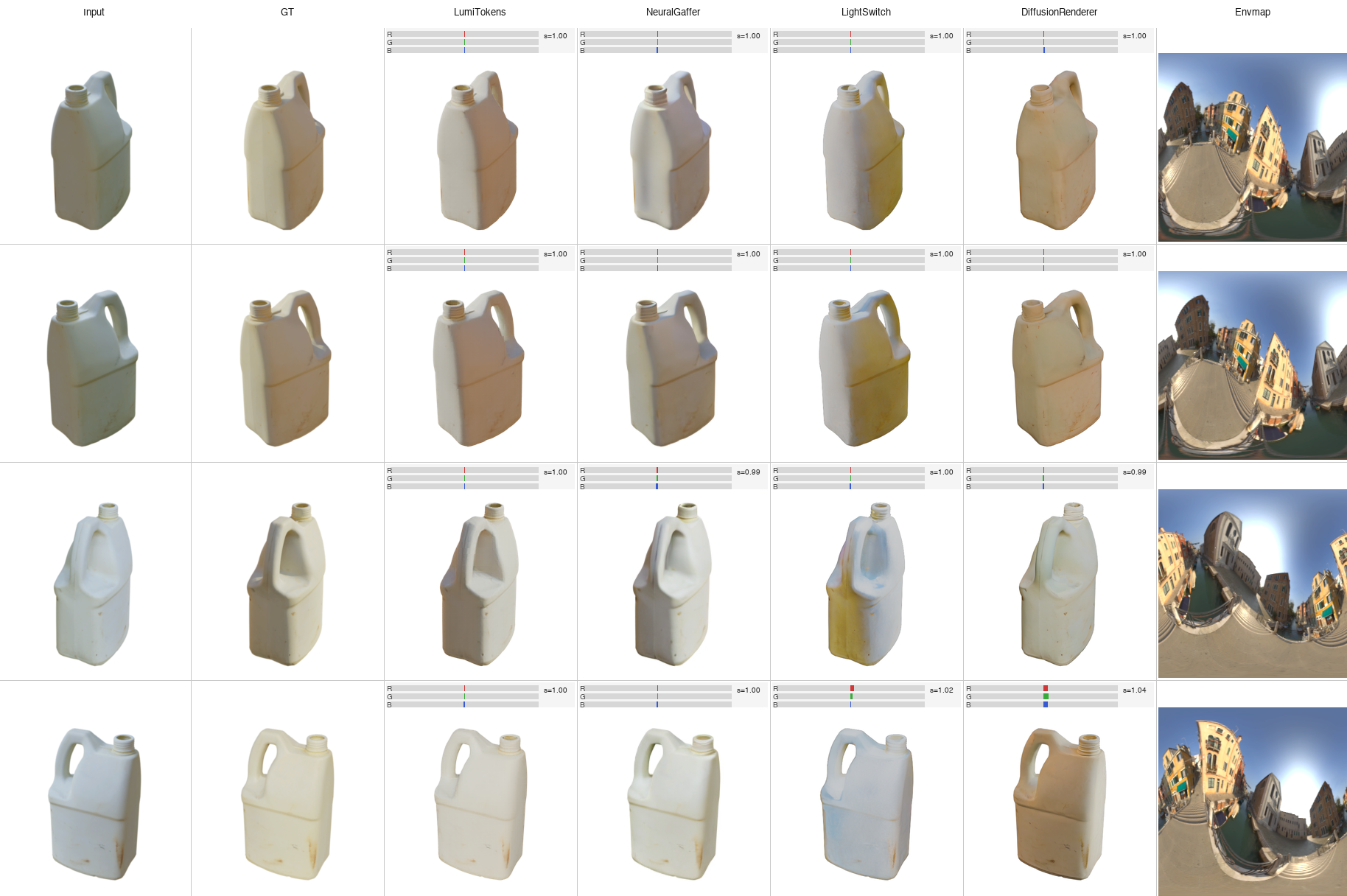}
        \subcaption{Plastic bottle gallon.}
    \end{minipage}
    \caption{Multi-view relighting comparisons. Per-view optimal tone scales (shown as sliders above each image) reveal that LumiToken exhibits more stable scaling across views than baselines, demonstrating better cross-view consistency.}
    \label{fig:supp:mv_relight_comparisons_part2}
\end{figure}

We include additional qualitative comparisons for multi-view relighting results in Fig.~\ref{fig:supp:mv_relight_comparisons_part1} and Fig.~\ref{fig:supp:mv_relight_comparisons_part2}. We show the images before tone-scaling and visualize the optimal tone scales to apply to the RGB channels as sliding bars above the images. The scales that need to be applied to LumiTokens are much more stable than others, demonstrating better consistency across views.

\begin{figure}[t]
    \centering
    \includegraphics[width=1.0\linewidth]{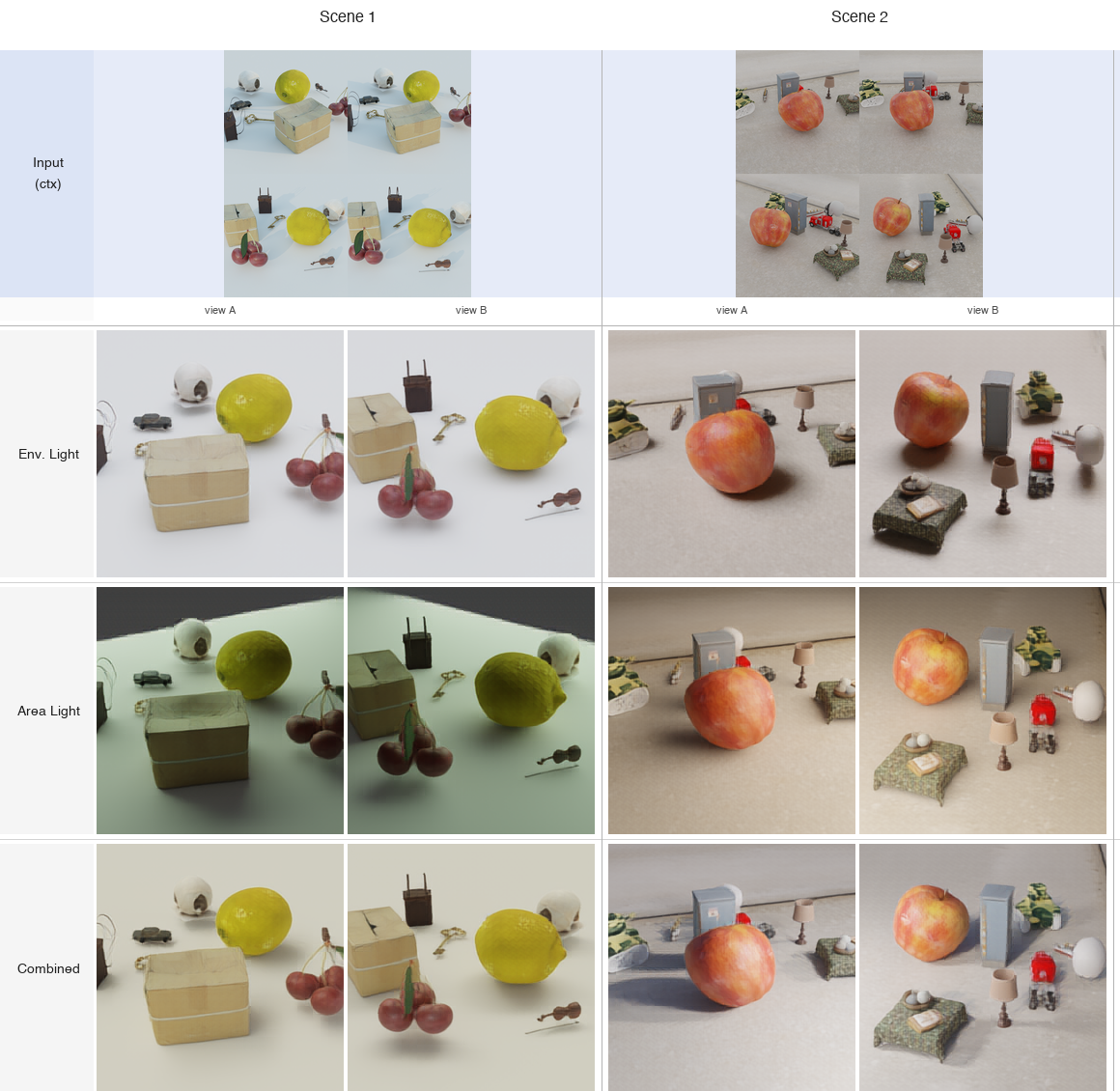}
    \caption{Qualitative relighting results on synthetic table-top scenes.}
    \label{fig:supp:scene_relighting_12}
\end{figure}

\begin{figure}[t]
    \centering
    \includegraphics[width=1.0\linewidth]{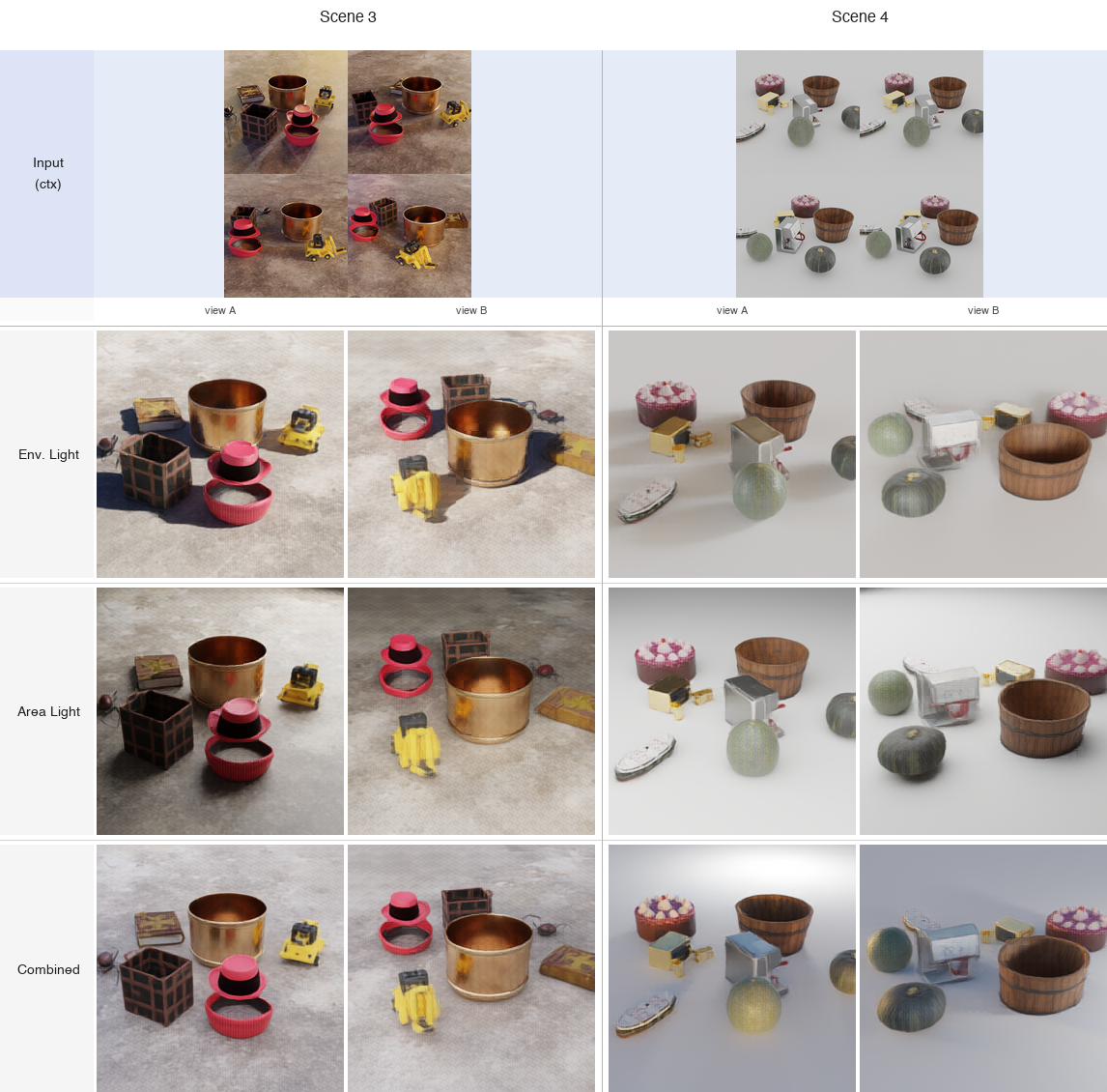}
    \caption{Qualitative relighting results on synthetic table-top scenes (continued).}
    \label{fig:supp:scene_relighting_34}
\end{figure}

We include additional qualitative results for relighting on synthetic table-top scenes in Fig.~\ref{fig:supp:scene_relighting_12} and Fig.~\ref{fig:supp:scene_relighting_34}.

\begin{figure}[t]
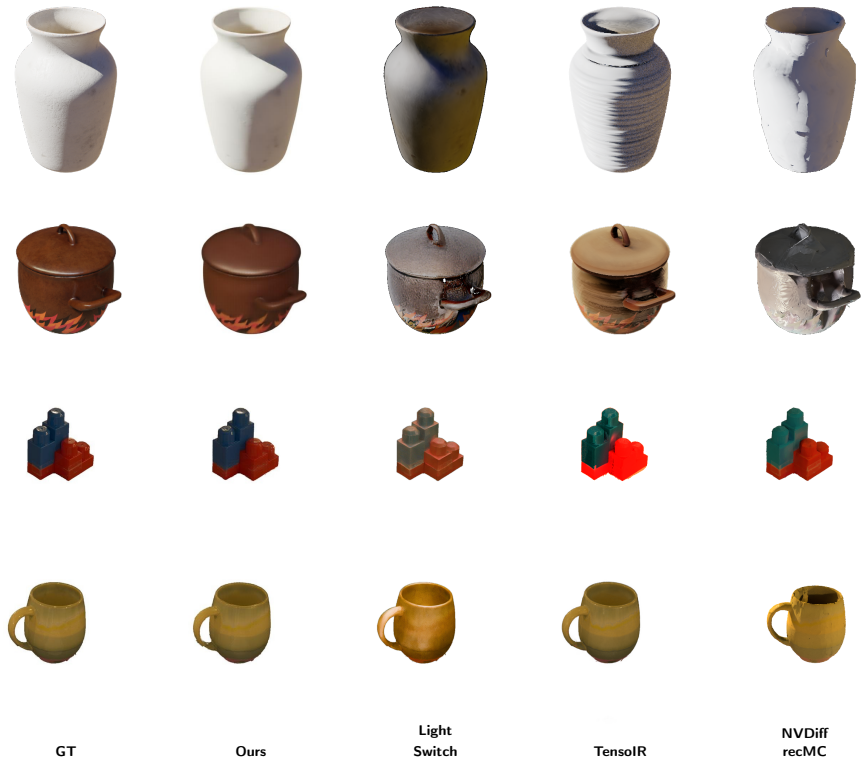

  \centering
  \setlength{\tabcolsep}{0.6pt}%
  \renewcommand{\arraystretch}{0.0}%
  \newcommand{\figfiveimg}[1]{%
    \adjincludegraphics[trim={0} {0.03\height} {0} {0.03\height},clip,width=0.187\linewidth,height=0.187\linewidth]{figures_reOrganized/fig5/#1}%
  }%
  \newcommand{\figfiveimgB}[1]{%
    \adjincludegraphics[trim={0} {0.03\height} {0} {0.03\height},clip,width=0.187\linewidth,height=0.187\linewidth]{figures_reOrganized/fig5_additional/#1}%
  }%
  \newcommand{\figfivelab}[1]{\parbox[t]{0.187\linewidth}{\centering\tiny\sffamily\bfseries\shortstack{#1}}}%
  \resizebox{\linewidth}{!}{%
  \begin{tabular}{ccccc}
    \figfiveimg{r1c1} & \figfiveimg{r1c2} & \figfiveimg{r1c3} & \figfiveimg{r1c4} & \figfiveimg{r1c5} \\
    \figfiveimg{r3c1} & \figfiveimg{r3c2} & \figfiveimg{r3x3} & \figfiveimg{r3c4} & \figfiveimg{r3c5} \\
    \figfiveimgB{sto_r2c1} & \figfiveimgB{sto_r2c2} & \figfiveimgB{sto_r4c3} & \figfiveimgB{sto_r2c4} & \figfiveimgB{sto_r2c5} \\
    \figfiveimgB{sto_r4c1} & \figfiveimgB{sto_r4c2} & \figfiveimgB{sto_r2c3} & \figfiveimgB{sto_r4c4} & \figfiveimgB{sto_r4c5} \\
    \figfivelab{GT} & \figfivelab{Ours} & \figfivelab{Light\\Switch} & \figfivelab{TensoIR} & \figfivelab{NVDiff\\recMC} \\
  \end{tabular}%
  }
  \caption{\textbf{Visual comparison of novel-view relighting.} The first two rows are synthetic experiments; the last two rows are real-world examples from Stanford-ORB.}
  \label{fig:supp:nv_relight_additional}
\end{figure}

We provide additional novel-view relighting comparisons in Fig.~\ref{fig:supp:nv_relight_additional}, following the same setup as in the main paper. The first two rows show synthetic experiments, while the last two rows show real-world examples from Stanford-ORB.

\end{document}